\documentclass[review,3p]{elsarticle}
\usepackage{multirow}
\usepackage{subfig}
\usepackage{url}

\usepackage{microtype}
\usepackage{amsmath}
\usepackage{cases}
\usepackage{algorithm}
\usepackage{algorithmicx}
\usepackage{algpseudocode}
\usepackage{fixltx2e}
\usepackage{graphicx}
\usepackage{graphicx,epstopdf}
\usepackage{array}
\usepackage{slashbox}

\usepackage{amssymb}

\journal{Image and Vision Computing}

\begin{document}

\begin{frontmatter}



\title{A Study on Wrist Identification for Forensic Investigation}


\author{Wojciech Michal Matkowski}
\author{Frodo Kin Sun Chan}
\author{Adams Wai Kin Kong}

\address{School of Computer Science and Engineering, Nanyang Technological University, Block N4, Nanyang Avenue, Singapore 639798}

\address{}

\begin{abstract}
Criminal and victim identification based on crime scene images is an important part of forensic investigation. Criminals usually avoid identification by covering their faces and tattoos in the evidence images, which are taken in uncontrolled environments. Existing identification methods, which make use of biometric traits, such as vein, skin mark, height, skin color, weight, race, etc., are considered for solving this problem. The soft biometric traits, including skin color, gender, height, weight and race, provide useful information but not distinctive enough. Veins and skin marks are limited to high resolution images and some body sites may neither have enough skin marks nor clear veins. Terrorists and rioters tend to expose their wrists in a gesture of triumph, greeting or salute, while paedophiles usually show them when touching victims. However, wrists were neglected by the biometric community for forensic applications. In this paper, a wrist identification algorithm, which includes skin segmentation, key point localization, image to template alignment, large feature set extraction, and classification, is proposed. The proposed algorithm is evaluated on NTU-Wrist-Image-Database-v1, which consists of 3945 images from 731 different wrists, including 205 pairs of wrist images collected from the Internet, taken under uneven illuminations with different poses and resolutions. The experimental results show that wrist is a useful clue for criminal and victim identification.
\end{abstract}


\end{frontmatter}

\clearpage
\section{Introduction}
\begin{figure}[!t]
\centering
\includegraphics[width=5.5in]{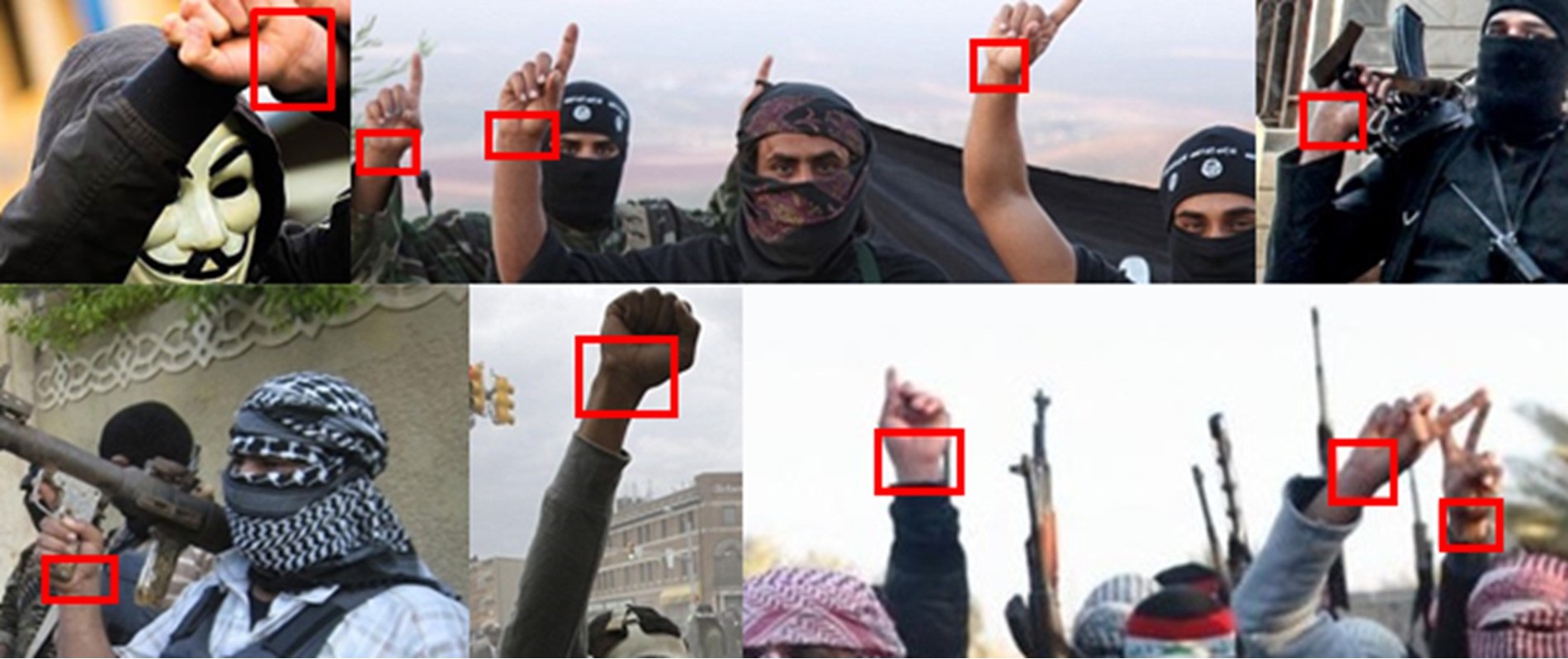}
\caption{Examples of images containing rioters and terrorists exposing their wrists. Child pornographic images also contain paedophiles' wrists but they cannot be put here because showing and processing them is illegal.}
\label{figT}
\end{figure}

Advances in technology have led to use digital evidence in forensic investigations and courts \cite{JonathanW.Hak2003}. In some cases, digital images may be the only available evidence which allows identifying criminals or victims. The evidence images, especially in the cases of child sexual abuses, riots, and terrorist activities, are usually taken with a high quality camera or mobile phone and photographed by criminal themselves, their partners or reporters. Some of these images are uploaded on the Internet websites called Clearnet by forensic investigators or the Dark web, including Freenet, I2P, and Tor, which allows users to be anonymous. The Dark web is a big source of crime scene images for forensic investigation \cite{AndyGreenberg2014}.\\
\indent 
Terrorists and rioters usually cover their faces with masks or clothes, but at the same time they may expose their forearms and hands, e.g. to express a triumph, greeting and salute or holding weapons etc. Images including rioters usually have higher resolution than terrorist images, if they are taken by press photographers such as the case of the masked Baltimore rioter who was photographed more than twice by Associated Press photographer exposing his wrist \cite{AshleyCollman2015} (see Fig. \ref{figWristInternetSamples}d). Terrorist images are more challenging because they often are low resolution. These images are captured for communication, threatening, or propaganda purpose. \\
\indent   
Another serious issue is child pornography, whose official term in the Interpol is child sexual abuse materials. The National Center for Missing \& Exploited Children (NCMEC) has revived more than 164 million abusive images and videos leading to identification of more than 10,900 child victims since 2002 \cite{NEMEC}. Assaulters take pictures for sale, personal records or exchange with other assaulters. A lot of these deals take place in the Dark web - study finds that over 80\% visits are related to child pornography \cite{AndyGreenberg2014}. To avoid recognition, abusers hide or blur their faces and tattoos in the abusive videos and images. However, other body parts such as back, chest, thigh, arm or hand are still visible. Some of these images can be close-up images with good quality and relatively high resolution.\\
\indent 
Identification from images becomes very challenging if there are no obvious characteristics available like faces or tattoos. Some of the recent studies have searched for new biometric traits, such as vein, skin marks, androgenic hair, and hand's victory sign patterns \cite{Zhang2012,Nurhudatiana2015,Chan2015,Hassanat2016}, for tackling the scenario without faces or tattoos available. Existing palmprint recognition methods \cite{ZhenanSun2005,Wu2014,Kang2014,Fei2019} for matching non-latent palmprint images taken from digital cameras focus on commercial biometric application with user cooperation and well-controlled imaging environments rather than forensic application.  In addition to palmprint, other similar hand-based biometrics proposed in the literature include, e.g., finger surface \cite{Woodard2005} and finger-knuckle-print \cite{Zhang2010} in well-controlled and cooperative enviroments. Vein recognition is mainly used for commercial application, where vein images are captured under infrared light in well-controlled environments \cite{Kang2014}. Recently, researchers show that some veins are hidden in color images and can be visualized for forensic investigation \cite{Zhang2012}. The vein recognition methods for forensic application rely on high image resolution and clear visualized veins, which are in some cases difficult to be extracted due to high concentration of fat or melanin or low image quality \cite{Tang2011}. Skin marks are applicable to high resolution images only and androgenic hair does not always occur in some body parts e.g., wrists. Victory sign, palmprint and finger knuckle are not always observed in evidence images. On the other hand, wrist may be most likely to be observed even though the subjects are wearing long sleeves, holding weapons, posing a gesture of triumph, greeting or salute or touching victims in child sexual offense cases. However, according to our best knowledge, no one studied wrist identification for forensic application before. Fig. \ref{figT} and Fig. \ref{figWristInternetSamples} show some wrist images obtained from the Internet.\\
\indent The rest of this paper is organized as follows. In Section \ref{NTUdatabase}, a testing database NTU-Wrist-Image-Database-v1 is presented. In Section \ref{WMFAAalgorithm}, the proposed algorithm is provided in details. In Section \ref{experimentalResults}, different experiments are performed to evaluate the proposed algorithm. In Section \ref{discussion}, discussion is given. In Section \ref{conclusion}, the conclusion is given.

\section{The NTU Wrist Image Database}\label{NTUdatabase}
\begin{figure}[]
\centering
\subfloat[]{\includegraphics[width=5in]{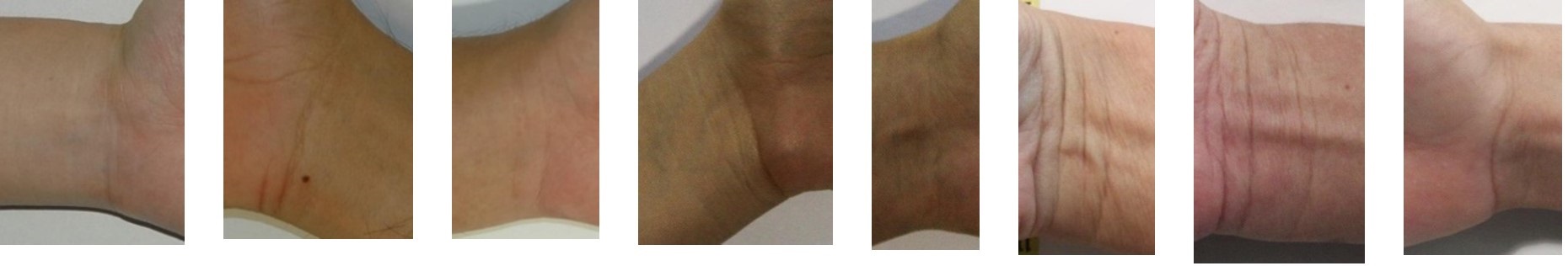}
\label{fig1dbStandardSet1}}
\hfil
\subfloat[]{\includegraphics[width=5in]{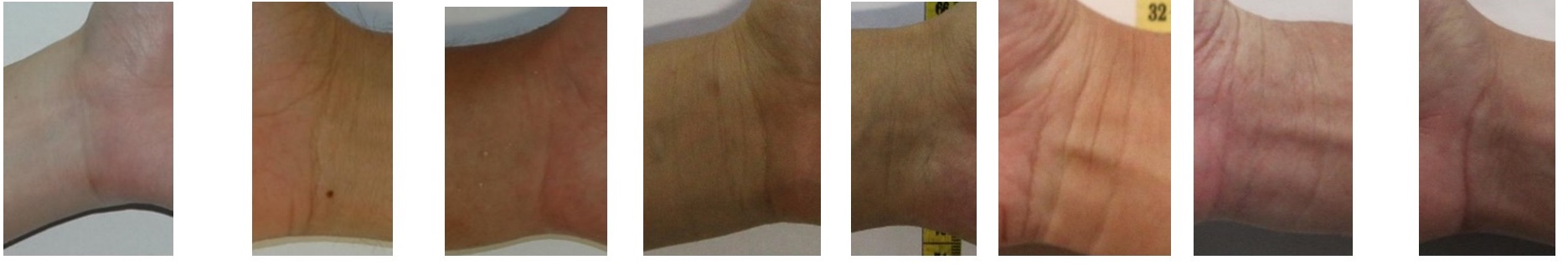}
\label{fig1dbStandardSet2}}
\hfil
\subfloat[]{\includegraphics[width=5in]{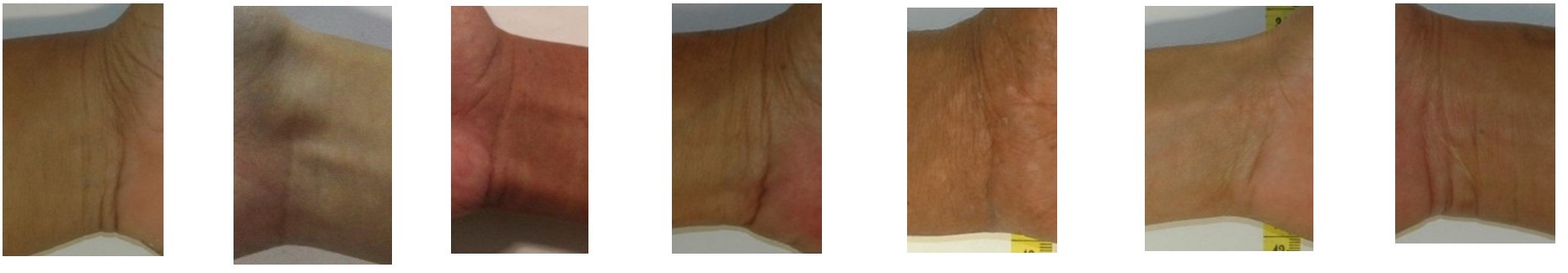}
\label{fig2dbLowQualitySet1}}
\hfil
\subfloat[]{\includegraphics[width=5in]{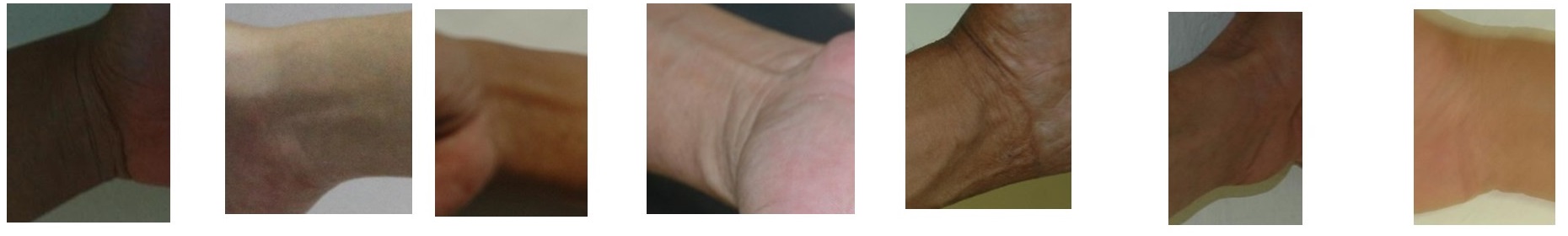}
\label{fig2dbLowQualitySet3}}
\hfil
\subfloat[]{\includegraphics[scale=1]{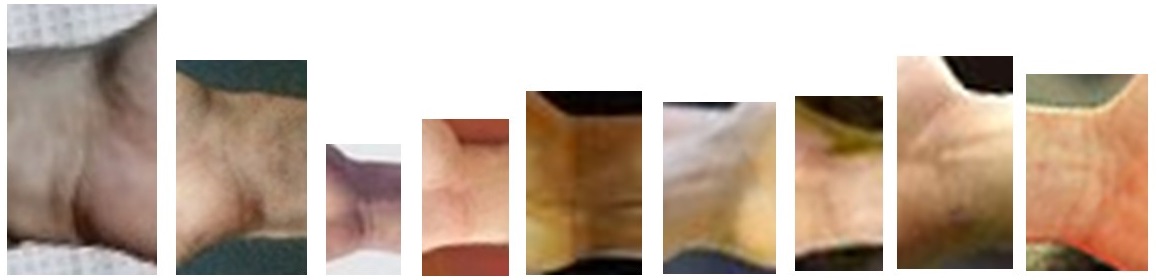}
\label{fig3dbInternetSet4}}
\hfil
\subfloat[]{\includegraphics[scale=1]{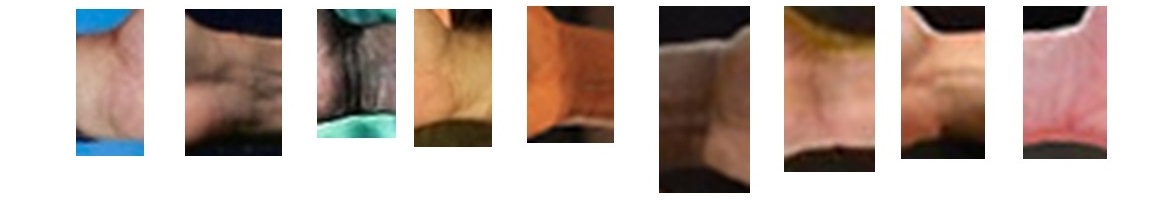}
\label{fig3dbInternetSet5}}
\caption{Examples of corresponding wrist images in the NTU-Wrist-Image-Database-v1: (a) and (c) Standard pose gallery images, (b) Standard pose probe images, (d) Non-standard pose probe images, (e) Internet gallery images and (f) Internet probe images. Original scales are preserved in (e) and (f). Images in (a), (b), (c) and (d) were resized down for better visualization. (a) and (b), (c) and (d), (e) and (f) show image pairs from the same writs.}
\label{figWristSamples}
\end{figure}

\begin{figure*}[!h]
\subfloat[]{\includegraphics[width=3in]{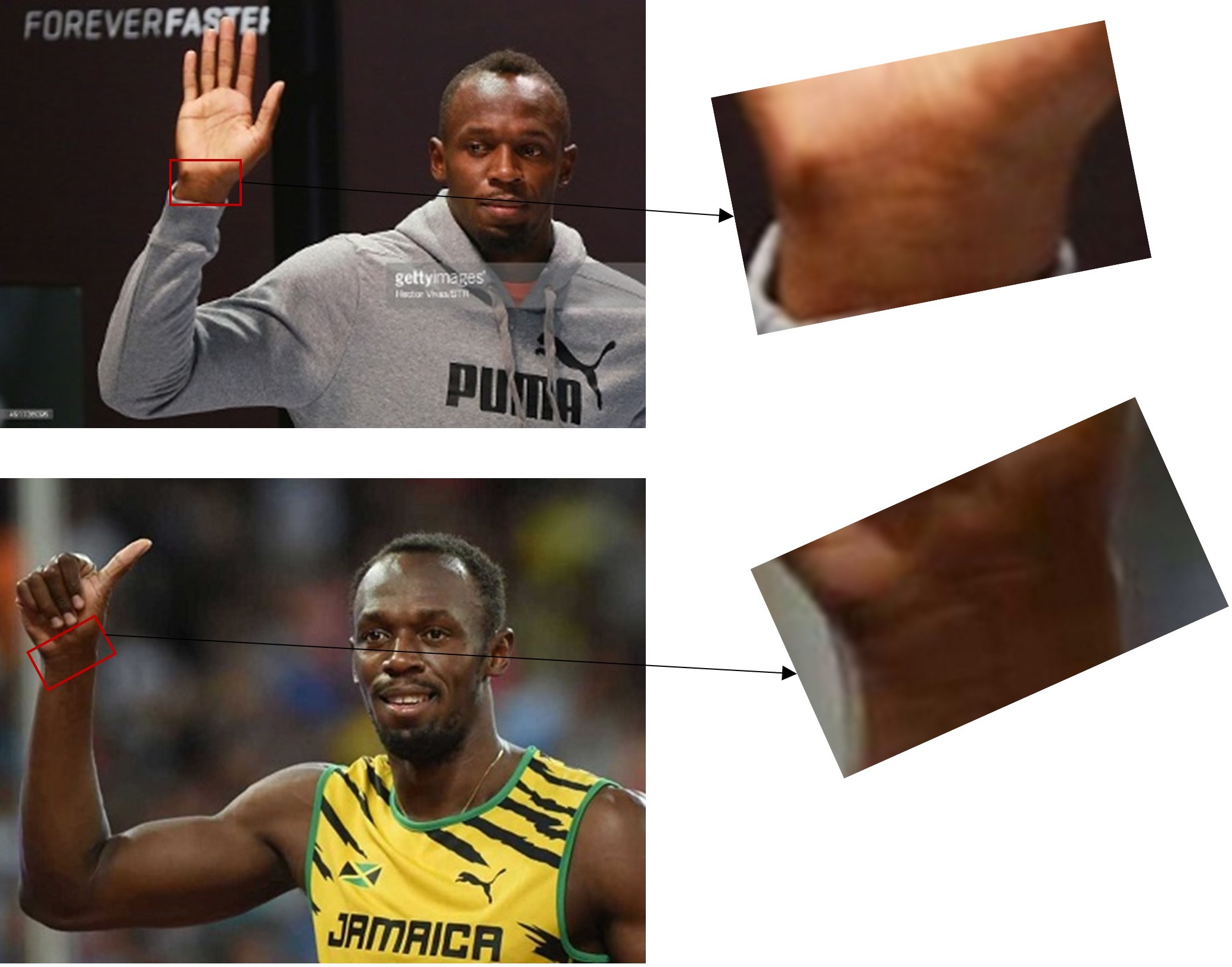}
\label{fig2samplesInternet}}
\hfil
\subfloat[]{\includegraphics[width=3in]{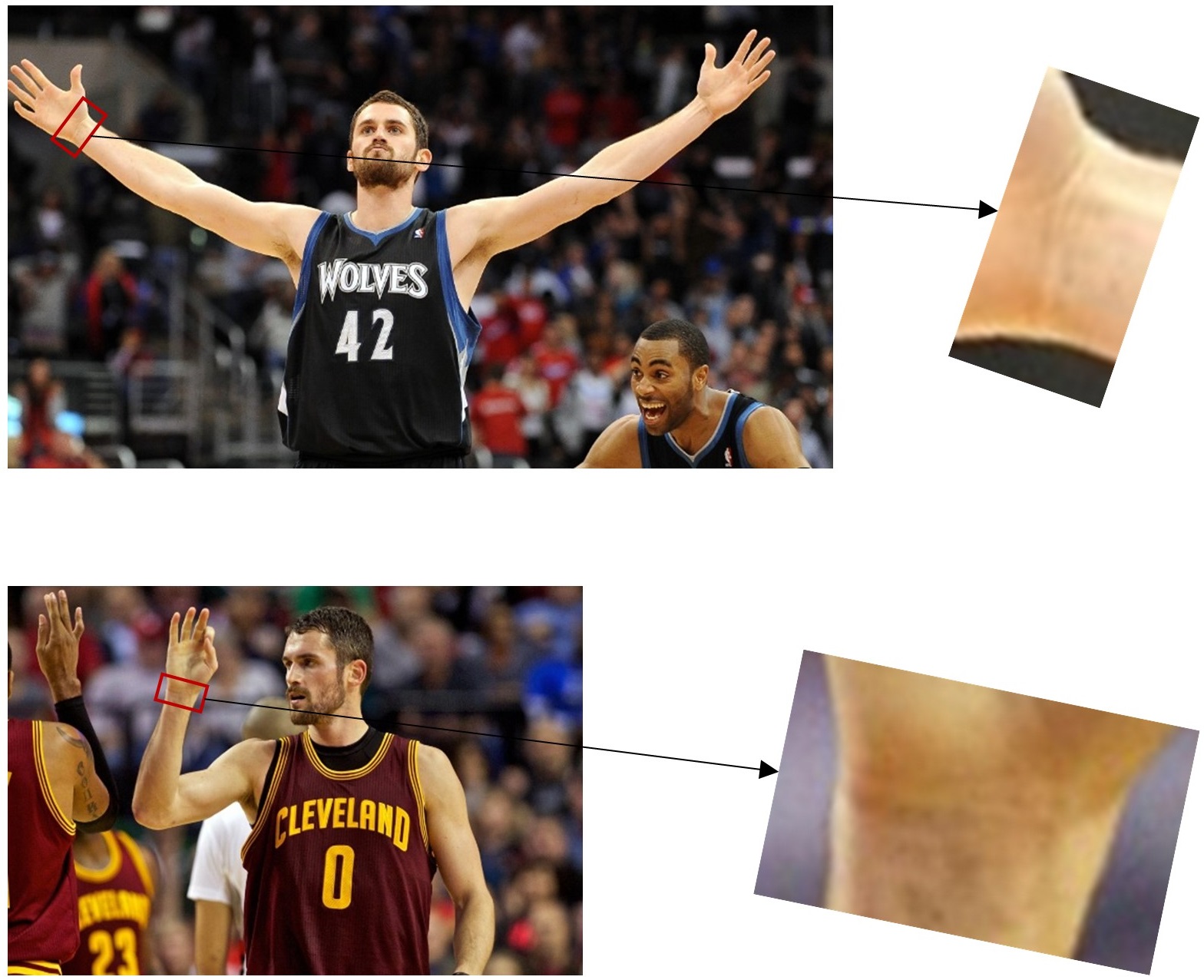}
\label{fig3samplesInternet}}
\hfil
\subfloat[]{\includegraphics[width=3in]{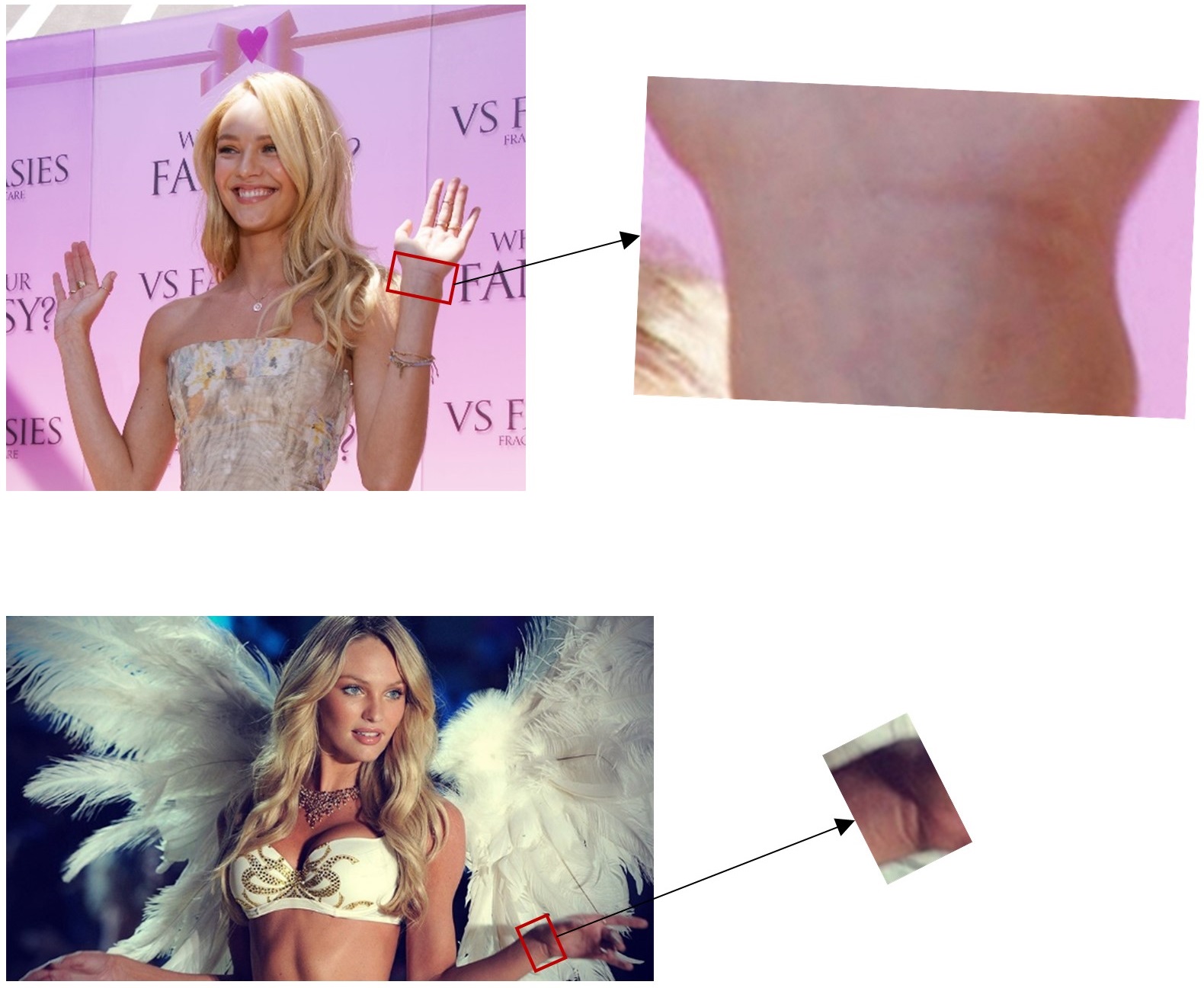}
\label{fig4samplesInternet}}
\hfil
\subfloat[]{\includegraphics[width=3.3in]{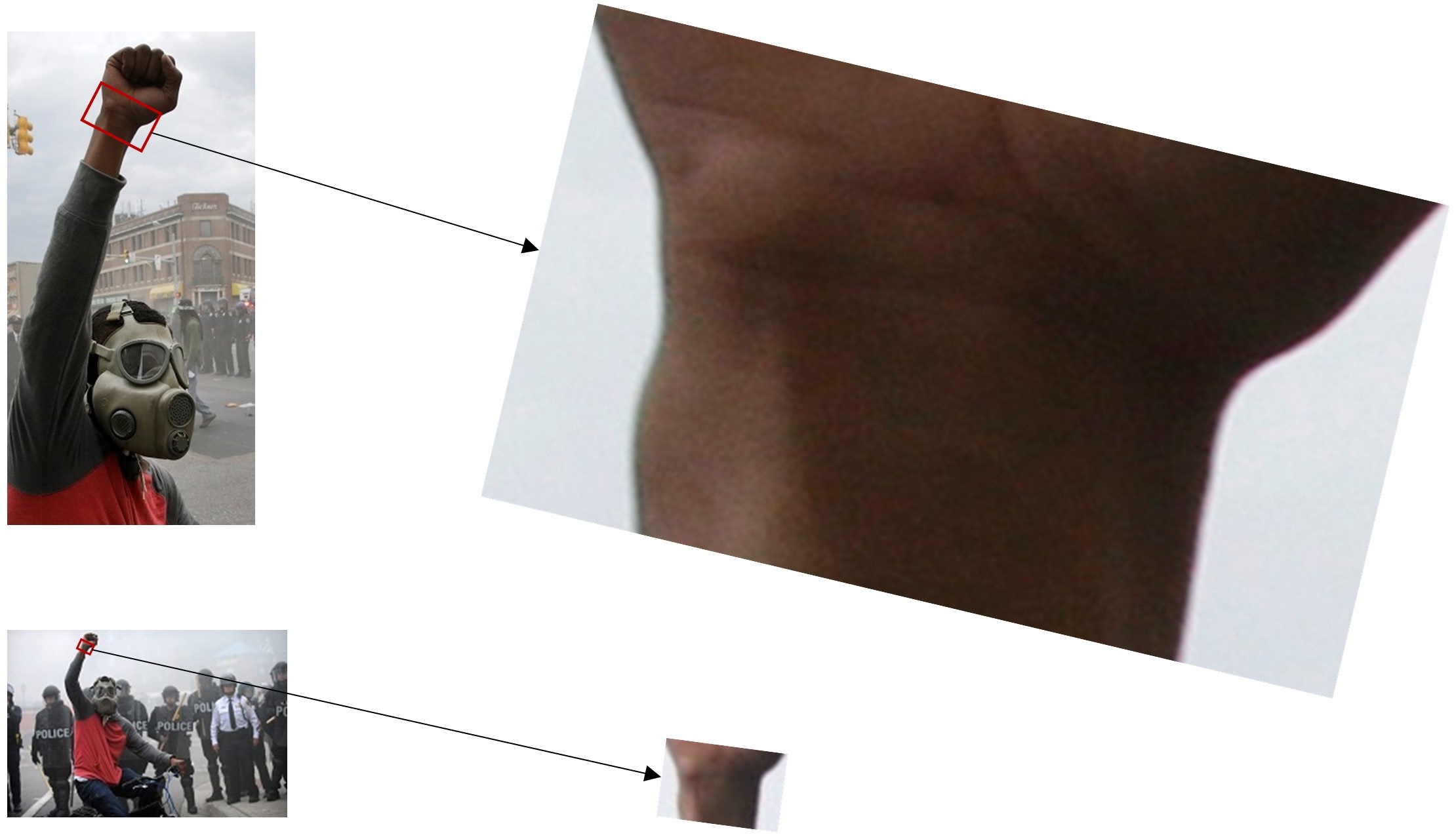}
\label{fig5samplesInternet}}
\caption{Examples of resized down original Internet images from the NTU-Wrist-Image-Database-v1. Wrists are highlighted in the original images, cropped and enlarged for better visualization. The scale is preserved between corresponding wrist images.}
\label{figWristInternetSamples}
\end{figure*}

Wrist images collected in Singapore were taken during two occasions with more than one week time interval. The images were cropped such that nothing but wrist area from the palmar side is visible\footnote{Note, that the wrist detection is out of scope in this study. There are some studies focusing on body parts detection in controlled \cite{Huynh2014} or uncontrolled environments \cite{Oliveira2016}. However, this work aims to show that wrist is a useful clue for identification.}. The widths and heights of the wrist images vary and on average they have 525 by 320 pixels. They are from left and right wrists of Asians - Chinese, Indian, Malay and also some Caucasian and Eurasian subjects. To increase the number of images in experiments, right hand wrists images were flipped such that they emulate left hand wrists from different subjects. The dataset was divided into three exclusive subsets. Gallery set SET1 includes 1948 images from 526 different wrists from 320 subjects taken during the first session. Probe set SET2 includes 1452 images from 397 different wrists from 282 subjects, taken during the second session. In both sessions, wrist images were taken in a frontal (palmar) pose without strict pose requirements and they are considered as standard pose images. Non-standard set SET3 includes 135 images from 133 different wrists of 84 subjects taken in varying conditions with viewpoint, pose and out of focus (OOF) variations. Images in SET1, SET2 and SET3 were taken by Canon EOS 500D or NIKON D70s cameras. To evaluate the proposed algorithm in more challenging cases, which are more similar to forensic environment, two additional datasets SET4 and SET5 were built. Gallery SET4 includes all images from SET1 and 205 additional images downloaded from the Internet (SET1P). Probe set SET5 contains 205 corresponding Internet wrist images. The wrists in the Internet images usually have low resolution and their sizes vary from 21 by 12 to 196 by 126 pixels. Most of them have less than 100 by 50 pixels. Some examples of the standard, non-standard pose and Internet images are shown in Fig. \ref{figWristSamples} and the dataset details are given in Tables \ref{tabDB} and \ref{tabDB2}. SET4 and SET5 are omitted in Table \ref{tabDB2} because SET4 is SET1 with additional 205 different wrist images and SET5 contains corresponding 205 wrist images. The NTU-Wrist-Image-Database-v1 will be available online \cite{ntuDatabase} for research purpose in 3 months after this paper is published.

\begin{table}[!h]
\renewcommand{\arraystretch}{1.1}
\caption{Details of the datasets in the NTU-Wrist-Image-Database-v1 \cite{ntuDatabase}.}
\label{tabDB}
\centering
\begin{tabular}{c|c|c|c|c}
\hline
\bfseries Set & \bfseries No.  &\bfseries No. & \bfseries No.  &\bfseries Description\\
& \bfseries images & \bfseries wrists & \bfseries subjects \\
\hline\hline
SET1 & 1948 & 526 & 320 & Gallery standard pose images \\
\hline
SET2 & 1452 & 397 & 282 & Probe standard pose images \\
\hline
SET3 & 135 & 133 & 84 & Non-standard images with illumination, pose, OOF variations\\
\hline
SET4 & 2153 & 731 & 505 & SET1 + Internet images SET1P (205 images from 205 wrists)\\
\hline
SET5 & 205 & 205 & 185 & Internet images \\
\hline
\end{tabular}
\end{table}
\normalsize

\begin{table}[!h]
\renewcommand{\arraystretch}{1.1}
\caption{Number of images for one wrist in SET1, SET2 and SET3.}
\label{tabDB2}
\centering
\begin{tabular}{c|c|c|c|c|c|c|c|c}
\hline
\backslashbox{\bfseries Wrists}{\bfseries Images} & \bfseries 1  &\bfseries 2 & \bfseries 3  &\bfseries 4
& \bfseries 5 & \bfseries 6 & \bfseries 7 & \bfseries 8\\
\hline\hline
SET1 & 6 & 86 & 37 & 361 & 5 & 29 & 0 & 2\\
\hline
SET2 & 4 & 62 & 23 & 295 & 5 & 7 & 0 & 1\\
\hline
SET3 & 131 & 2 & 0 & 0 & 0 & 0 & 0 & 0\\
\hline
\end{tabular}
\end{table}
\normalsize

\clearpage
\section{The Proposed Wrist Identification Algorithm}\label{WMFAAalgorithm}
The WMFA (Wrist Matcher for Forensic Applications) algorithm consists of four main steps. First, for image pre-processing, a skin segmentation step employing superpixels and ensemble of decision trees is applied (Section \ref{wristSemgentation}). Then, wrist region of interest (ROI) is found by using a ROI extraction scheme (Section \ref{roiExtraction}). Then, large feature sets are extracted (Section \ref{featureExtraction}). Next, a one against all classifier-based matching scheme is used (Section \ref{wristMatching}). On top of these, there is a post-recognition score analysis which aims to improve the overall accuracy by combining four WMFAs with different settings to build the whole WMM (Wrist Meta-Matching) system (Section \ref{metaRecognition}).

\subsection{Wrist Segmentation}\label{wristSemgentation}
\begin{figure}[]
\centering
\includegraphics[width=6in]{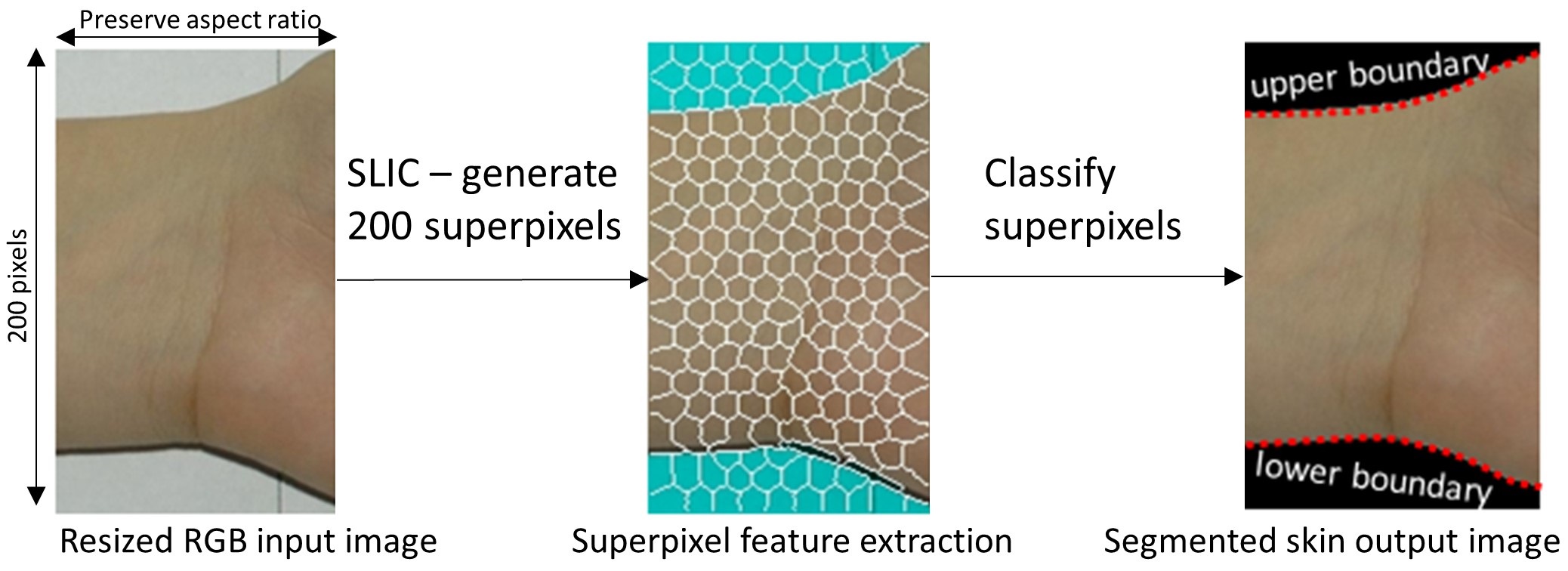}
\caption{The wrist segmentation scheme.}
\label{figSegmentation}
\end{figure}

The WMFA algorithm is fully automated except segmentation for non-standard images and does not require any manual and time consuming human supervision. For establishing large forensic datasets, automatization is required to effectively collect and process images from prisoners or suspects, but for crime scene images, manual correction is acceptable. Currently, law enforcement agencies also use a semi-automatic approach to handle latent prints. Since the region of interest (ROI) is a wrist, wrist skin pixels are segmented for further analysis. Segmentation based on a simple threshold may not be ideal because the skin of different population has diverse color. Even in the same wrist image, skin color can be different significantly because palms and upper wrist regions have less concentration of melanin, comparing with lower wrist regions and forearms, especially for brown, dark brown and black skin persons, i.e., Fitzpatrick scale V and VI. Spatial arrangement and patch based information can increase the accuracy of skin segmentation \cite{Kruppa2002} and therefore, superpixel methods, which group pixels into perceptually meaningful patches, are used. Simple linear iterative clustering (SLIC) is a popular and fast method that has desirable properties: superpixels adhere well to boundaries and they improve segmentation results \cite{Achanta2011}. Thus, SLIC is employed for this study. Before applying SLIC, images are resized such that their heights are fixed at 200 pixels and their widths are varied to preserve aspect ratio. Then, SLIC method is used to generate 200 superpixels for each image. Within each superpixel and 8 neighbors, mean and standard deviation (statistics) values from RGB, HSV, LAB, YCbCr, YIQ, normalized RGB color spaces and seven gradient maps (Sobel in two directions, Prewitt in two directions, Laplacian, Difference of Gaussians, Laplacian of Gaussians) are extracted to form 450-dimensional feature vectors\footnote{Dimension of one superpixel description: [6 color spaces] x [3 channels] x [2 statistics] + [7 gradient maps] x [2 statistics] = 50). Because each superpixel is represented by itself and 8 neighbors then the feature vector size of one superpixel is: 50 + 50 x 8 = 450.}. To segment skin superpixels, an ensemble of decision trees (EoDT) is trained with the bagging method. EoDT classifies each superpixel into either skin or non-skin class, resulting in a segmented wrist image. The wrist segmentation scheme is illustrated in Fig. \ref{figSegmentation}.

\subsection{ROI Extraction}\label{roiExtraction}
\begin{figure}[!h]
\centering 
\includegraphics[width=6in]{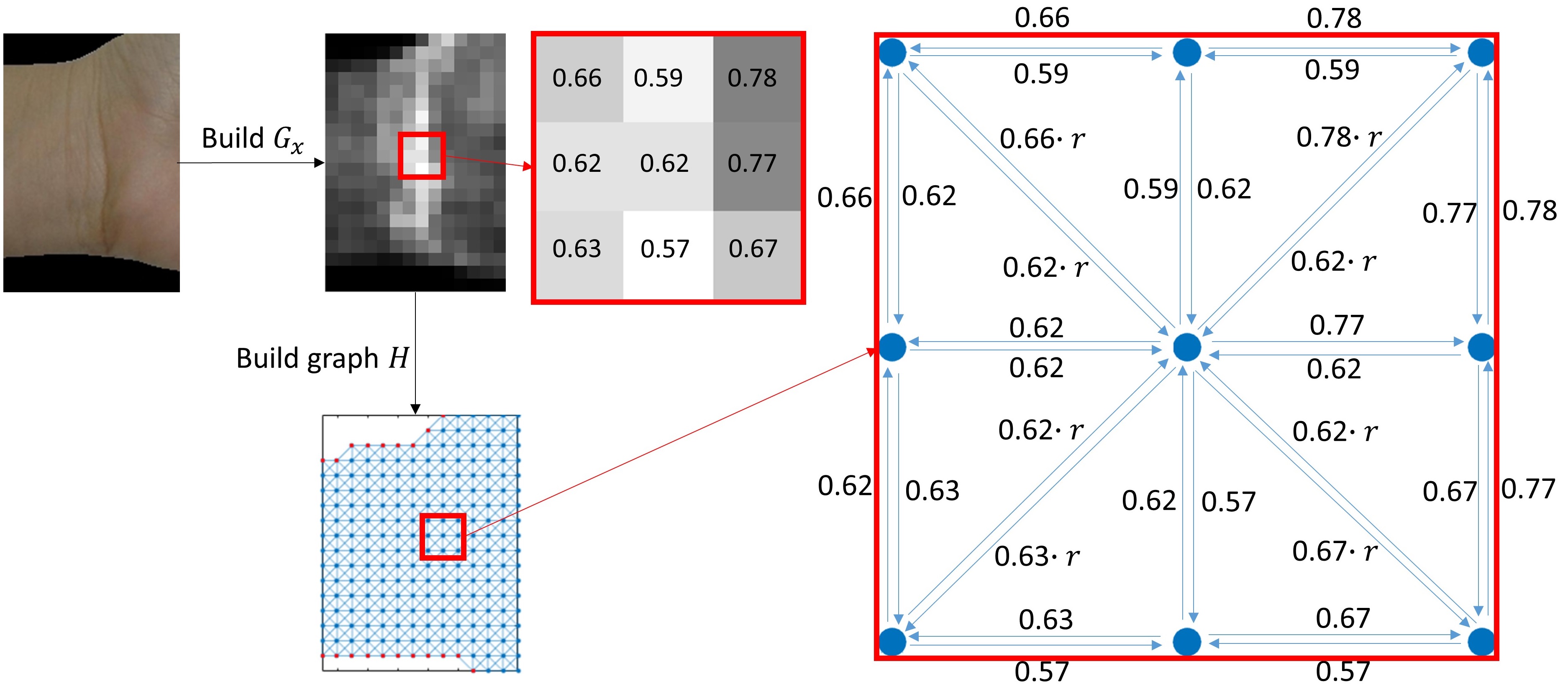}
\caption{Illustration of graph edges and weights between vertices determined based on the gradient image.}
\label{graph}
\end{figure}

\begin{figure}[]
\centering
\subfloat[]{\includegraphics[width=1in]{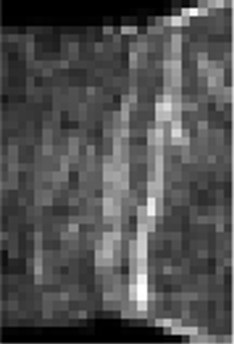}
\label{fig1wrinkles}}
\hfil
\subfloat[]{\includegraphics[width=1in]{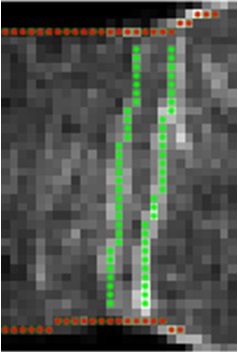}
\label{fig2wrinkles}}
\hfil
\subfloat[]{\includegraphics[width=1in]{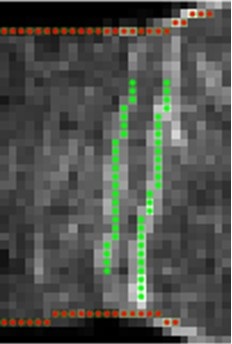}
\label{fig3wrinkles}}
\hfil
\subfloat[]{\includegraphics[width=1in]{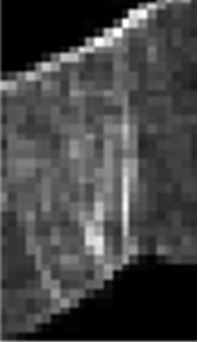}
\label{fig4wrinkles}}
\hfil
\subfloat[]{\includegraphics[width=1in]{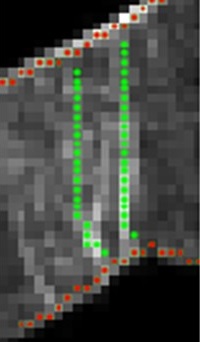}
\label{fig5wrinkles}}
\hfil
\subfloat[]{\includegraphics[width=1in]{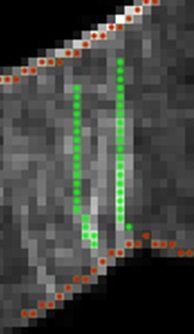}
\label{fig6wrinkles}}
\caption{(a), (d) gradient maps from two wrist images. (b), (e) key points extracted using Proc2 from (a), (d). (c), (f) key points extracted using Proc2/3 from (a), (d). Red dots are boundaries while green dots are extracted wrinkles.}
\label{figWrinklePoints}
\end{figure}

This section aims to extract a common ROI from the segmented wrists. At the beginning, it should be emphasized that the wrist area is not well defined. It is located in between hand and forearm. Human hand has distinctive and common key points, i.e. hand shape allowing efficient palm ROI extraction for commercial palmprint and palm vein applications \cite{Wu2014, Kang2014}. But this work is for forensics so most of the hand information is assumed not available which makes ROI extraction more challenging. In real applications, hands of criminals can be occluded by weapons, e.g., guns, when they hold them. Pose variation is caused by different posing among different subjects and scale variation is caused by uncontrolled imaging environments. \\
\indent 
To address these pose and scale issues, a robust ROI extraction scheme is developed. First, since the images are segmented in the previous section, lower and upper boundaries (see Fig. \ref{figSegmentation}), which will serve as key points, are easily extracted. Based on the observation of the wrists, they usually have around two prominent wrinkles. The ROI of the wrists is bounded by the lower and upper boundaries in the horizontal direction and has two wrinkles in the vertical direction (see Fig. \ref{figWrinklePoints}). To find ROI, a two stage scheme is proposed for detecting the wrinkle and boundary points. In the first stage, to extract vertical lines, the segmented wrist image channels $I_c$, where $c$ index denotes R, G and B channels of the segmented image $I$ are convolved with the Sobel operator in the vertical direction obtaining $G_{rx},G_{gx},G_{bx}$  respectively. The final gradient image is obtained by selecting maximum response for each pixel as follows $G_x= 1-max⁡(|G_{rx} |,|G_{gx} |,|G_{bx} |)$. This operation enhances vertical wrinkles as shown in Figs. \ref{fig1wrinkles} and \ref{fig4wrinkles}. Pre-processed image $G_x$ is resized to have 40-pixel height. Now, a small directed sparse graph $H=(V,E)$ is built with Moore neighbourhood connectivity (8 neighbours) within the wrist, which is actually a quasi-King's graph. The graph is based on the gradient image of wrist pixels. In other words, a graph node exists at some position $(i,j)$ if there is a skin pixel at that position represented by mask image $M(i,j)=1$. The distance to reach the node from its neighbour is equal to a gradient image value at the position. To regulate distances in diagonal directions in point 6 of the Algorithm \ref{buildGraph}, assuming that one pixel is a unit square and following Pythagorean Theorem, $r=\sqrt{2}$ is used. Fig. \ref{graph} illustrates the graph. This stage is described in detail in Algorithm \ref{buildGraph}.

\begin{algorithm}
\caption{Build graph}\label{buildGraph}
\begin{algorithmic}[1]
\Procedure{BuildGraph}{$I,M$}
\State $G_{cx}\gets\begin{bmatrix} -1 & 0 & 1 \\
-2 & 0 & 2 \\
-1 & 0 & 1
\end{bmatrix} \star I_c$ \Comment{Convolve images}
\State $G_x \gets 1 - max(|G_{rx}|,|G_{gx}|,|G_{bx}|)$
\State Resize $G_x$ to $40$ pixel height
\State $H=(V,E)$, it has vertices in the foreground: $V \in \{v_{i,j}|M(i,j)=1\}$. It has edges between eight \indent closest pixel neighbours: $E\{(v_{i,j},v_{i+m,j+n})\}$, $m,n=-1,0,1 \wedge m\ne n\ne 0$   \Comment{Define connections}
\State \[w(v_{i,j},v_{i+m,j+n})=\]
\[\begin{cases} G_x(i+m,j+n)\cdot r & |m|=|n| \wedge m\ne n\ne 0 \\
G_x(i+m,j+n) & otherwise \end{cases} \] \Comment{Define weights}
\State \textbf{return} $H$\Comment{Sparse graph $H$}
\EndProcedure
\end{algorithmic}
\end{algorithm}

Next stage aims to find two prominent wrinkles. Given that the graph represents an energy map, wrinkles are believed to be localized such that they contain minima points between two boundaries. By taking advantage of the small sparse graph representation generated by Algorithm \ref{buildGraph}, the shortest paths $P_1$ and $P_2$ between the upper boundary $b_{up}$ and the lower one $b_{down}$ are found using Johnson's algorithm, which searches the shortest paths between all pairs of vertices and is applicable for sparse graphs \cite{ThomasH.CormenCharlesE.Leiserson2009}. The upper boundary $b_{up}$ determines a set of start nodes $S$ and the lower boundary $b_{down}$ determines destination nodes $F$. The first shortest path is found in the original graph $H$. Since the second shortest path is expected not to contain the nodes in  $P_1$, $P_2$ is found in a subgraph. The subgraph is obtained by subtracting $P_1$ and its von Neumann neighbours (four adjacent nodes) from the graph $H$. This stage is described in Algorithm \ref{detectWrinkles}. Algorithm \ref{detectWrinkles} may be run with ($adjust$ in stage 4 and 10 of Algorithm \ref{detectWrinkles} is set to be true) or without the procedure described in Algorithm \ref{adjustPath} inside, which are denoted respectively by Proc2/3 and Proc2. Proc2 has a hard constraint on the start point and destination point which have to be localized on boundaries $b_{up}$ and $b_{down}$. However, observing wrist wrinkles, one can say that they are not always situated from boundary to boundary. To increase the representational power and make the algorithm more robust, Algorithm \ref{adjustPath} is also proposed. Proc2/3 employing both Algorithms \ref{detectWrinkles} and \ref{adjustPath} aims to relax the constraint by introducing the normalized path and searching for the minimal one between shifted down the start nodes and shifted up the destination nodes which are controlled by a parameter $a$. Parameter $a$ determines the length of the wrinkles, such that higher $a$ results in shorter wrinkles. The difference between wrinkle-like key points returned by Proc2 and Proc2/3 is presented in Figs. \ref{figWrinklePoints} and \ref{figRoiExtraction}.

\begin{algorithm}
\caption{Detect wrinkles}\label{detectWrinkles}
\begin{algorithmic}[1]
\Procedure{DetectWrinkles}{$H,b_{up},b_{down},a,adjust$}
\State $S \in \{s_i|s_i \in b_{up}\}, F \in \{f_j|f_j \in b_{down}\}$\Comment{Define nodes based on boundaries}
\State Apply Johnson's algorithm on $H$ to get all shortest paths $P_{All}$
\State Among all the shortest paths $P_{All}$ find the shortest one $P_1$ between $S$ and $F$
\If{$adjust$ = true}
\State $P_1\gets$\Call{AdjustPath}{$P_1,P_{All},b_{up},b_{down},a$}\Comment{Run Algorithm \ref{adjustPath}}
\EndIf
\State $V_{sub} \gets ((V-P_1)\ominus B)\cup S \cup F$
\State $H = (V_{sub},E)$ \Comment{Graph modification}
\State  Apply Johnson's algorithm on $H$ to get all shortest paths $P_{All}$ 
\State Among all the shortest paths $P_{All}$ find the shortest one $P_2$ between $S$ and $F$
\If{$adjust$ = true}
\State $P_2\gets$\Call{AdjustPath}{$P_2,P_{All},b_{up},b_{down},a$}\Comment{Run Algorithm \ref{adjustPath}}
\EndIf
\State \textbf{return} $P_1,P_2$\Comment{Two wrinkles}
\EndProcedure
\end{algorithmic}
\end{algorithm}

\begin{algorithm}
\caption{Adjust the path}\label{adjustPath}
\begin{algorithmic}[1]
\Procedure{AdjustPath}{$P_a,P_{All},b_{up},b_{down},a$}
\State $d_p \gets \frac{1}{n}\sum_{v \in P_a} w(v_k,v_{k+1})$, $k=1,2,...,n+1.$ $\#P_a=n+1$ \Comment{Define normalized path length}
\For{$i \gets 1$ to $\lfloor n \cdot a \rfloor$} 
\For{$j \gets 1$ to $\lfloor n \cdot a \rfloor$}
\State Among all the shortest paths $P_{All}$ find the shortest one $P_s$ between $b'_{up,i}$ and $b'_{down,j}$
\If{$\frac{1}{n-i-j}\sum_{v \in P_s} w(v_k,v_{k+1}) < d_p$, $k=i+1,...,n-j$}
\State $d_p \gets \frac{1}{n-i-j}\sum_{v \in P_s} w(v_k,v_{k+1})$
\State $P_a \gets (v_{i+1},...,v_{n-j})$ 
\EndIf
\EndFor
\EndFor
\State \textbf{return} $P_a$\Comment{Adjusted path}
\EndProcedure
\end{algorithmic}
\end{algorithm}

\begin{figure}[]
\centering
\subfloat[]{\includegraphics[width=2in]{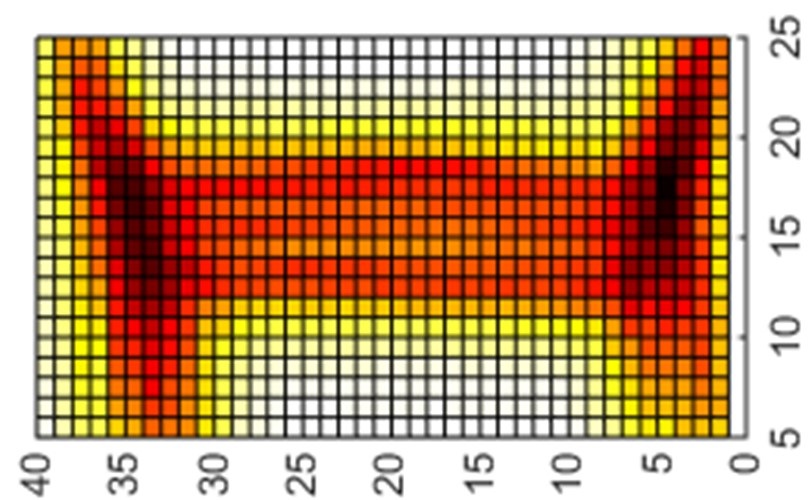}
\label{figHeatMap}}
\hfil
\subfloat[]{\includegraphics[width=2in]{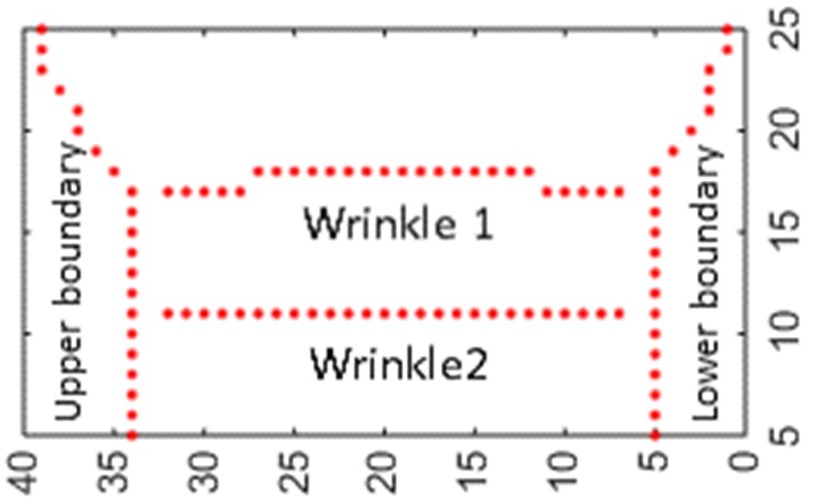}
\label{figTemplatePoints}}
\hfil
\caption{(a) Heat map representing sum of the key points from SET1 images. (b) wrist template is built by taking the reference of the heat map. Both figures are rotated 90$^\circ$ counter-clockwise.}
\label{figTemplate}
\end{figure}

\begin{figure}[!t]
\centering
\includegraphics[width=6in]{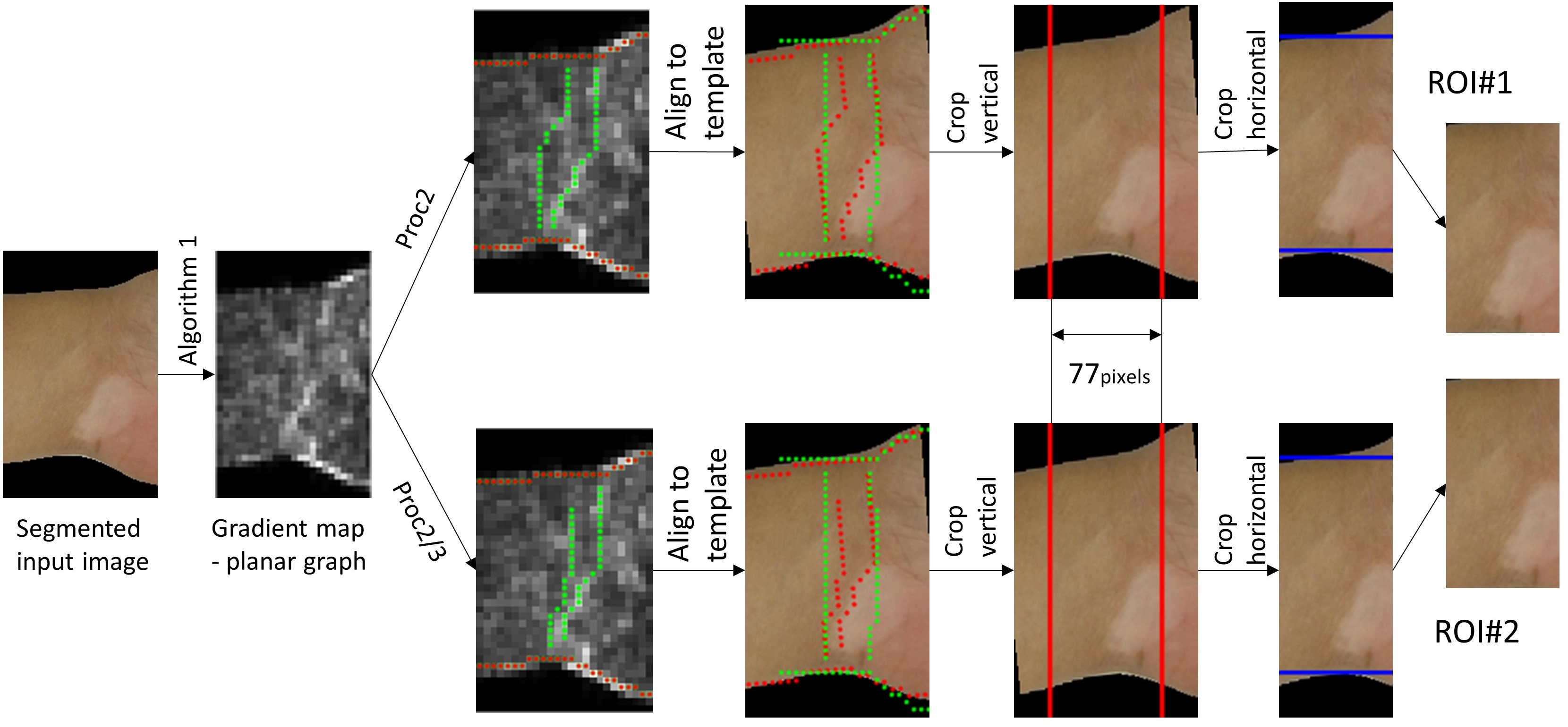}
\caption{ROI extraction steps using both Proc2 and Proc2/3 in order to produce respectively ROI\#1 and ROI\#2.}
\label{figRoiExtraction}
\end{figure}

In the Procedures, a structuring element $B$ is set up to be a 3x3 cross. Parameter $a$ is set up to $0.2$. Operator $\ominus$ denotes morphological erosion, $\cup$ is a union of sets, $\#$ is the number of elements in the set and $\lfloor \rfloor$ is an operator of a floor function. Points $b'_{up,i}$ and $b'_{down,j}$ are start and destination nodes shifted in vertical direction by respectively $i$ pixels down and $j$ pixels up. Using Proc2, a wrist template is estimated based on all extracted key points from the training images in SET1. More precisely, a heat map is obtained by finding the boundaries $b_{up}$ and $b_{down}$ and the two vertical wrinkles $P_1$ and $P_2$ in each training image and summing them up. Using the heat map (Fig. \ref{figHeatMap}) as a reference, template (Fig. \ref{figTemplatePoints}) for ROI extraction is proposed. By representing the template consisting of extracted wrist boundaries and wrinkles as 2D points, Coherent Point Drift (CPD) registration method, which fits the Gaussian mixture model centroids from one set to another set of points by maximizing likelihood, is employed to find the correspondence between template key points and wrist key points in input images under affine transformation. Detailed description of the CPD method can be found in \cite{Myronenko2010}. After applying CPD to align images to the template, a ROI is found by cropping columns of pixels from both, left and right sides of the image. In the next step, a ROI is constrained by two horizontal lines, whose positions are defined based on the number of foreground pixels in a row. If more than three quarters of pixels in a row are foreground pixels, then the entire row is considered as a foreground, otherwise a background. As it was mentioned before, there are two procedures Proc2 and Proc2/3 that produce slightly different key points and therefore different alignments and ROIs are expected. ROI\#1 and ROI\#2 denote the ROIs extracted by Proc2 and Proc2/3, respectively. These two procedures relax the assumptions on the wrist wrinkles and provide more flexible representation because each wrist can be now represented by ROI\#1 and ROI\#2. Extracted key points are shown in Figs. \ref{fig2wrinkles}, \ref{fig3wrinkles}, \ref{fig5wrinkles}, \ref{fig6wrinkles}. The ROI extraction step is illustrated in Fig. \ref{figRoiExtraction}, which also presents how one can obtain different ROIs using different key point localization procedures. 

\subsection{Feature Extraction}\label{featureExtraction}
After image alignment is performed and ROI is obtained, features, including local binary patterns (LBP) in each RGB channel, Gabor orientation field histograms and dense scale invariant feature transform (DSIFT) on a grayscale image, are extracted. These features have been successfully applied to biometric applications \cite{Ahonen2006,Su2014,Wang2010}. LBP features, including $LBP_{D,R}^{riu2}$   and uniform $LBP_{D,R}^{u2}$, are useful for texture recognition. $LBP_{D,R}^{riu2}$  is invariant to rotation and monotonic grey level changes. Uniform $LBP_{D,R}^{u2}$ has at most two bitwise transitions from 0 to 1 or 1 to 0 when considered as circular patterns and is invariant to monotonic grey level changes \cite{Ojala2002}. Gabor filters capture line segments \cite{Su2014} and DSIFT describes shape related information \cite{Wang2010}. Raw $LBP_{D,R}$ is obtained as:
\begin{equation}
\label{eqlbp}
LBP_{D,R}=\sum_{p=0}^{D-1}\xi(g_p-g_c)2^p
\end{equation}

\begin{figure}[]
\centering
\subfloat[]{\includegraphics[scale=1]{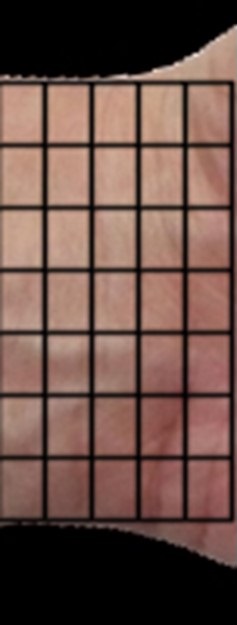}
\label{fig1grids}}
\hfil
\subfloat[]{\includegraphics[scale=1]{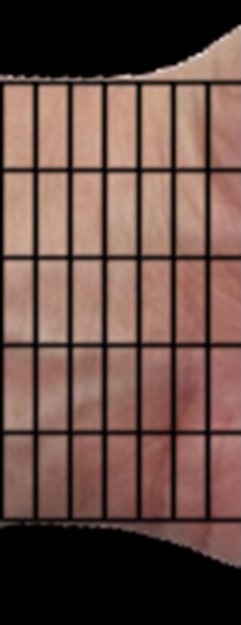}
\label{fig1grids2}}
\hfil
\subfloat[]{\includegraphics[scale=1]{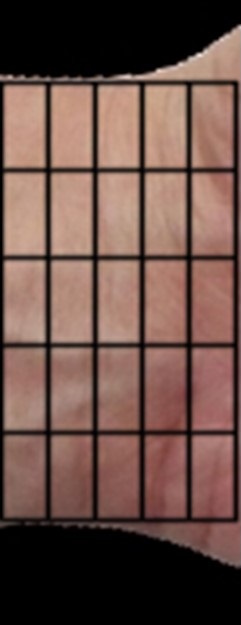}
\label{fig2grids}}
\hfil
\subfloat[]{\includegraphics[scale=1]{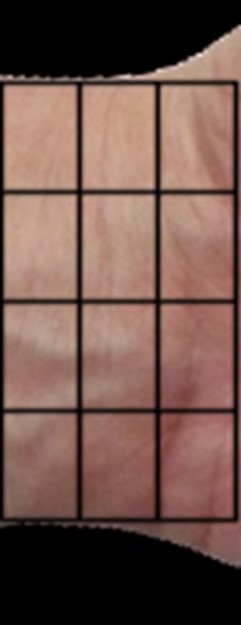}
\label{fig2grids2}}
\hfil
\subfloat[]{\includegraphics[scale=1]{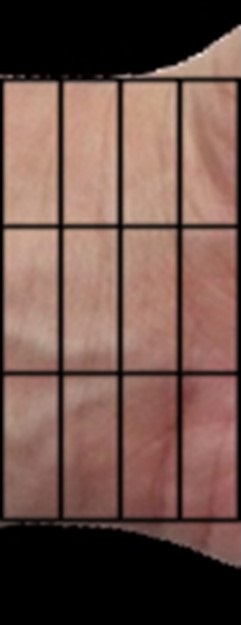}
\label{fig3grids}}
\hfil
\subfloat[]{\includegraphics[scale=1]{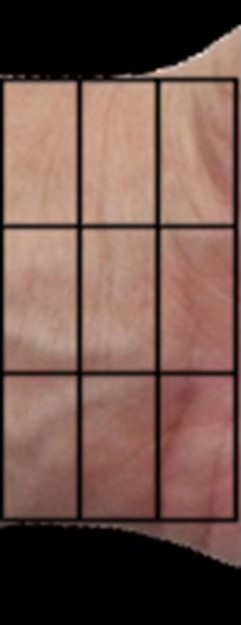}
\label{fig3grids2}}
\hfil
\subfloat[]{\includegraphics[scale=1]{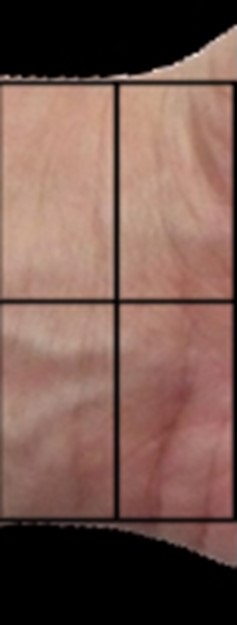}
\label{fig4grids}}
\caption{Seven different grids (a) $b_1$, (b) $b_2$, (c) $b_3$,(d) $b_4$, (e) $b_5$, (f) $b_6$, (g) $b_7$.}
\label{figGrids}
\end{figure}

\begin{subnumcases}{\xi(g_p-g_c)=}
1 & $g_p \geq g_c $\\
0 & $g_p < g_c$
\end{subnumcases}
where $D$ is the number of sampling points, $R$ is the radius of sampling circle, $g_c$ is the centre pixel value and $g_p$ is its neighbour pixel value. Gabor orientation field features are defined the same as in \cite{Su2014}. These features are obtained by convolving image $I$ with Gabor filters $G$ as follows:
\begin{equation}
O(x,y)=\arg_{\theta_k}\underset{m,k}{\max}|G_{\lambda_{mk},\theta_l,\sigma_m,\lambda} \star I|
\end{equation}
where $G_{\lambda_{mk},\theta_l,\sigma_m,\lambda}$ is a Gabor filter with an orientation $\theta_k=k\pi/8$, sinusoidal component of the wavelength $\lambda_{mk}$, standard deviation of the elliptical Gaussian window $\sigma_m$, spatial aspect ratio $\lambda$, scale indices $m$ and orientation indices $k$. DSIFT features are extracted by a SIFT descriptor on a regular grid controlled by pixel step, spatial bin size and area where it should be extracted. Detailed SIFT description can be found in \cite{Lowe2004}. To preserve spatial information, ROI is divided into many blocks forming a grid structure. Seven different grids $b_1$, $b_2$, $b_3$, $b_4$, $b_5$, $b_6$, $b_7$ shown in Fig. \ref{figGrids} are proposed to extract Gabor and LBP histograms from each block. Each DSIFT feature is extracted from a specified fixed block. First, a part of the feature vector $f_{LBP}$ is extracted on R, G and B channels and consists of the concentrated histograms from  $LBP_{8,1}^{riu2}$ on $b_1$, $b_2$ and $LBP_{8,2}^{u2}$ on $b_3$, $b_4$, $b_5$, $b_6$, $b_7$. Two different LBP features are used because applying  $LBP_{8,2}^{u2}$ on the first two grids which have small blocks would make the histogram sparse. On the other hand, applying rotational invariant $LBP_{8,1}^{riu2}$ on bigger blocks would cause the loss of discriminative directional information in already aligned images. Gabor filters with 16 different orientations and 4 scales $s\in \{0.2,0.5,0.7,0.9\}$ on all grids are used to extract the second part of the feature vector $f_{Gabor}$. Moreover, DSIFT features, which are extracted with 16 pixel step and spatial bin size of 16 are added as feature vector $f_{DSIFT}$. Feature vectors $f_{LBP},f_{Gabor},f_{DSIFT}$ dimensions are 13074, 2112 and 1280 respectively. The final 16466-dimensional feature vector $f=\{f_{LBP},f_{Gabor},f_{DSIFT}\}$ is used to represent each ROI of the wrist image. 

\subsection{Wrist Matching}\label{wristMatching}
Partial least squares (PLS) regression is a statistical method that employs latent variables to find the relations between two mean centred and variance scaled matrices $X$ and $y$. In this study, $y$ is in fact a vector. It uses the properties of nonlinear iterative partial least squares (NIPALS) algorithm to calculate the principal components iteratively \cite{Geladi1986}. PLS decomposes matrices $X$ and $y$ into: 
\begin{equation}
X = TP^\top+E
\end{equation}
\begin{equation}
y=UQ^\top+F
\end{equation}
where $T$ and $U$ are score matrices, $P$ and $Q$ are loading matrices and $F$ and $E$ are residual matrices and $^\top$ denotes transpose operator. In order to build classifiers, the regression coefficients $\beta$ are found such that matching score can be obtained by $\hat y = \beta^\top x+\bar y$ where $\hat y$ ̂  is a classifier response or matching score, $x$ is an input feature column vector and $\bar y$ is the sample mean. Detailed description of PLS and the computation of $\beta$ coefficients using NIPALS algorithm can be found in \cite{Geladi1986} and \cite{Rosipal2006} respectively. \\
\indent
Wrist identification can be performed as a multiclass classification. One-against-all classifiers are trained, which means that every wrist in the gallery set has its own unique classifier. All images in the gallery set are used to build the classifiers. In the training, label $y_i=1$ when the images belong to a particular wrist and $y_i=-1$ otherwise. $y_i$ is the $i\textsuperscript{th}$ element of $y$. In the experiments, PLS with $k=5$ is used and compared to Support Vector Machine (SVM). 

\subsection{Meta-Recognition}\label{metaRecognition}
\begin{figure}[]
\centering
\includegraphics[width=5in]{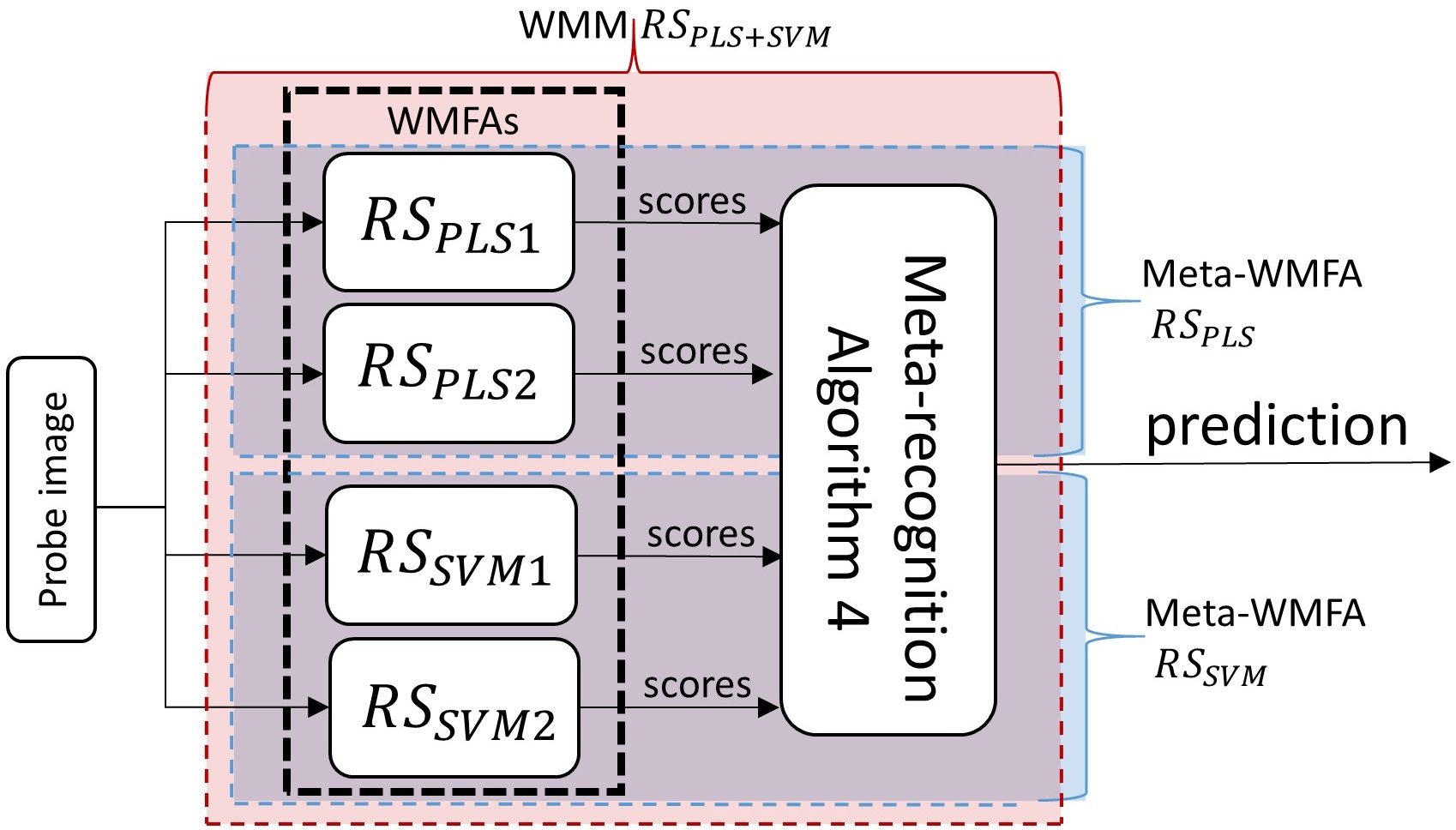}
\caption{The schematic diagram of the whole Wrist Meta Matching system (WMM) and other Meta-WMFA systems. Each $RS$ block represents a WMFA algorithm.}
\label{figMetaRecognition}
\end{figure}

To boost the matching performance by taking advantage of the two ROIs, a Meta-recognition approach is used. Here, the notion of a recognition system refers to the WMFA algorithm under a specific set of parameters described in Section \ref{roiExtraction} while the notion of Meta-recognition refers to analyse matching scores to select the best system for each input wrist.  Because there are two different procedures, Proc2 and Proc2/3 for ROI extraction, each image after alignment has ROI\#1 and ROI\#2, which are represented by two different feature vectors. It allows building two different PLS as well as two different SVM classifiers. In practice, it means that each gallery wrist has four different classifiers with input features extracted from the two ROIs.  In other words, there are four different recognition systems $RS_{PLS1}$, $RS_{PLS2}$, $RS_{SVM1}$, and $RS_{SVM2}$. Differences in particular steps in the recognition systems are presented in Table \ref{tabMeta}. \\
\indent
The Meta-recognition step analyses matching scores from each of the recognition systems and decides which one of them is the most likely correct. Scheirer et al. proposed a Meta-recognition system based on Extreme Value Theorem (EVT) also known as the Fisher-Tippet Theorem \cite{Scheirer2011}. The key idea behind assumes that there is a sufficient number of nonmatch samples in order to model the non-matched score distribution and the matched score is regarded as an outlier of the distribution. They also showed that modelling the nonmatch scores in a tail is an extreme value problem. Thus, the tail of the distribution can be modelled by one of extreme value distributions. According to \cite{Scheirer2011}, Weibull distribution is the most suitable one for modelling nonmatch distribution in statistical Meta-recognition. Its cumulative distribution function is expressed by:  $F(x)=1 - \exp({-(\frac{x}{a})^b})$ for $x \geq 0 $ and $F(x)=0$ for $x < 0 $.
where $b$ is a shape parameter and $a$ is a scale parameter. In the experiment, there are always $m-1$ nonmatch responses from $m$ different wrist classifiers and one match response from the correct one. Since it satisfies the assumption for modelling nonmatch distribution, Weibull-Based Statistical Meta-Recognition is adapted to run on top of our four different $RS_i$. In addition, there are three Meta-WMFA recognition systems for further comparison: $RS_{PLS}$, $RS_{SVM}$ and $RS_{(PLS+SVM)}=WMM$ which details are presented in Table \ref{tabMeta}. System $RS_{PLS}$ refers to WMFA algorithms with PLS-based matching step; $RS_{SVM}$ refers to the WMFA algorithms with SVM-based matching step and WMM is the Meta-recognition system that considers all combinations of the WMFA algorithms. A schematic diagram of the systems is presented in Fig. \ref{figMetaRecognition}. To choose the best $RS_{best}$ from four $RS_i$ for each input wrist image, we propose a modification of Rank-1 Statistical Meta-Recognition Algorithm from \cite{Scheirer2011} as shown in Algorithm \ref{weibull}. It fits Weibull distribution to $l_t-1$ highest matching scores from each recognition system skipping the top matching score $\hat y_{i,1}$ and determinates its parameters. Then, from the cumulative distribution function (CDF) $F_i$, one can answer the question what it the probability that the top matching score does not come from the distribution. Based on that, the preferable $RS_{best}$ would be the one with the highest value of its CDF calculated at the top matching score $\hat y_{best,1}$. Thus, the final prediction scores $s$ of the system are the scores returned by $RS_{best}$. In experiments, the length of the tail $l_t$ is experimentally set up to be a function of the gallery size $l_t=0.5 \cdot m$, where $m$ is the number of the classifiers in the gallery.

\begin{algorithm}
\caption{Modified Rank-1 Statistical Meta-Recognition}\label{weibull}
\begin{algorithmic}[1]
\Procedure{MetaRecognition}{$\hat Y_i$} \Comment{Sorted in descending order scores $\hat Y_i=(\hat y_{i,1},...,\hat y_{i,m})$ from $n$ recognition systems $RS_1,...,RS_n \in RS$, where $1 \leq i \leq n$}
\For{$i \gets 1$ to $n$}
\State Fit Weibull distribution $F_i$ to $l_t-1$ largest scores: $\hat y_{i,2},...,\hat y_{i,l_t}$
\EndFor
\State $k \gets \underset{i}{\mathrm{argmax}}[F_i(y_{i,1})]$ \Comment{Find $RS_{best}$ index $k$}
\State \textbf{return} $s \gets \hat Y_k$\Comment{Matching scores}
\EndProcedure
\end{algorithmic}
\end{algorithm}

\begin{table}[]

\renewcommand{\arraystretch}{1.1}
\caption{Recognition and Meta recognition systems details.}
\label{tabMeta}
\centering
\begin{tabular}{c|c|c|c|c}
\hline
\bfseries Meta-WMFA & \bfseries Meta-WMFA &\bfseries WMFA &\bfseries ROI & \bfseries Matching\\
\hline\hline
\multirow{4}{9em}{WMM\\$RS_{PLS+SVM}$} & 
\multirow{2}{4em}{$RS_{PLS}$} & $RS_{PLS1}$ & ROI\#1 & PLS \\
\cline{3-5}
&&$RS_{PLS2}$ & ROI\#2 & PLS \\
\cline{2-5}
&\multirow{2}{4em}{$RS_{SVM}$} & $RS_{SVM1}$ & ROI\#1 & SVM \\
\cline{3-5}
&&$RS_{SVM2}$ & ROI\#2 & SVM \\
\hline
\end{tabular}
\end{table}

\section{Experimental Results}\label{experimentalResults}
The WMFA algorithms and Meta-WMFA systems were evaluated under several aspects and compared with six state-of-the-art biometrics methods designed for palmprint, palm vein or fingerprint matching, Sun et al.'s OrdinalCode, Fei et al.'s double-orientation code (DOC), Wu et al.'s SIFT-based method, Kang et al.'s Mutual Foreground (MF)-Based LBP with $\chi^2$ distance (MF\_LBP), Minaee et al.'s Deep Scattering Convolutional Network (DSCN) and Translation Invariant Scattering Network (TISN). The methods were selected for the following reasons. 1) According to the best of our knowledge, this is the first work on wrist identification based on color images. 2) Palmprint nearby the wrist also has wrinkles and seems to be more similar than other body parts. 3) Sun et al.'s Ordinal Code \cite{ZhenanSun2005} is a well-known method which bases on image filtering and uses Hamming distance as a matching metric. 4) Fei et al.'s method achieves the highest accuracy among coding based approaches in palmprint verification and identification \cite{Fei2016}. 5) Wu et al.'s SIFT-based method \cite{Wu2014} is applicable to contactless palmprint verification and does not require contact image acquisition devices, which has some similarity to our study. 6) Kang et al.'s Mutual Foreground (MF)-Based LBP \cite{Kang2014} with $\chi^2$ distance is developed for contactless palm veins. It uses a mutual foreground between two images which is extracted based on maximal principal curvatures and similarly to WMFA algorithm, LBP is employed as features. 7) Mianee et al.'s TISN and DSCN are methods which employ deep features - a similar feature representation as deep convolutional networks (DCN) and achieve very promising results using principal component analysis (PCA) and multiclass SVM in fingerprint and palmprint recognition \cite{Minaee2015,Minaee2017}. Note that although in forensic applications, imaging environments are also contactless, our study and the previous studies are under very different assumptions. In Wu et al. and Kang et al.'s studies, the contactless imaging environment, including illumination setting, is under control and users cooperate with the imaging devices. In forensic applications, imaging environments are out of control and there is no user cooperation. Thus, large pose and illumination variations are expectable. 
All the experiments were performed on the NTU-Wrist-Image-Database-v1 using different combinations of the sets. Three experiments EXP1, EXP2, and EXP3 are considered as a representation of different forensic cases. Description of these experiments is presented in Table \ref{tabExpScenarios}. Detailed quantitative description of the image sets is given in Tables \ref{tabDB} and \ref{tabDB2}. Note that the gallery (training set) and the probe (testing set) are exclusive sets of images and each probe wrist has its corresponding wrist in the gallery.  The four WMFA recognition systems, $RS_{PLS1}$, $RS_{PLS2}$, $RS_{SVM1}$ and $RS_{SVM2}$ and Meta-recognition systems, $RS_{PLS}$, $RS_{SVM}$ and WMM were compared and the contribution of LBP, Gabor and DSIFT features for the proposed WMFA algorithm was evaluated. An algorithm is considered to be a superior or better when it achieves higher rank-1 accuracy. The results of the experiments are presented as cumulative match characteristic (CMC) curves in Figs. \ref{figExp1}, \ref{figExp2}, \ref{figExp3}, summarized in Table \ref{tabResults} and described in Section \ref{WMeva}. Additionally, in Section \ref{PCA}, PCA and minimal-redundancy-maximal-relevance (mRMR) criterion \cite{Peng2005} are considered to reduce the high-dimensional feature vectors  and the results are reported in Tables \ref{tabResultsPCA} and \ref{tabResultsmRMR}.\\
\indent 
The features employed in our algorithm: LBP, Gabor and DSIFT are well known features, successfully applied in many domains, including biometrics \cite{Chan2015,Ahonen2006,Su2014,Wang2010,Chan2017a}. Recently researchers tend to move towards deep features, which are believed to be more powerful than hand crafted features. For example, in \cite{Minaee2015} and \cite{Minaee2017} the authors apply deep features to fingerprint and palmprint recognition achieving very good results in controlled environment. These deep features are obtained from the layers of the scattering network which is described in \cite{Bruna2013}. There are also some recent advancements in hand crafted features e.g. multichannel decoded local binary pattern (mdLBP)\cite{Dubey2014}, which can be considered to enhance matching performance. However, the aim and focus of this work are not to find the best representation for a wrist but to show that wrist is a useful clue for criminal and victim identification. Additional experiments are performed to compare LBP with mdLBP and the hand crafted features with the deep features. The results of the experiments are summarized in Table \ref{tabResultsFeatures} and described in Section \ref{expFeat}.\\
\indent
Before classifiers training and wrist matching were performed, standard pose images were segmented automatically, while the Internet images were segmented manually, which is acceptable in forensic cases. Two ensembles of 300 decision trees, which served as skin superpixel classifiers, were trained by using bagging method. There were over 500,000 skin superpixels and 125,000 non-skin superpixels used to build the two-class classifiers. 3400 standard pose images were divided into two separate folds $F_1$ and $F_2$, which were used to train classifiers $C_1$ and $C_2$. Then, the segmentation results were obtained using $C_1$ ($C_2$) to classify superpixels of $F_2$ ($F_1$). Next, the standard pose SET1 and SET2, the non-standard pose SET3 and the Internet images SET5 and SET1P were automatically aligned to the template and the proposed features were extracted from ROI\#1 and ROI\#2, such that each image was represented by two feature vectors. Segmentation and alignment results were checked and no obvious errors were noted. Finally, for each wrist in the gallery set, two PLS and two SVM classifiers were trained using one-against-all approach. 
Note that when applying the matching methods from \cite{ZhenanSun2005,Wu2014,Kang2014,Fei2016,Minaee2015,Minaee2017}, input ROI images are always extracted using our algorithms (see Sections \ref{wristSemgentation} and \ref{roiExtraction}), because none of these methods can handle the wrist segmentation, alignment and ROI extraction. Thus, the comparisons among WMM, WMFAs and other methods are at the feature and matching levels. Moreover, if there is no Meta-recognition step on the top of a matching algorithm and ROIs are not indicated, the results are always presented for the ROI that gives higher result in terms of rank-1 accuracy. \\
\indent
The proposed WMM algorithm is implemented in MATLAB and run on a PC with Xeon E5 3.5 GHz CPU. During the pre-processing stage, the average time of skin segmentation (Section \ref{wristSemgentation}) is 5 seconds/image, which gives 0.025 seconds for classification of one superpixel. Note that this step is run only once because after skin segmentation step, all segmented images are stored. For ROI extraction (Section \ref{roiExtraction}) and feature extraction (Section \ref{featureExtraction}), the average time cost is 0.03 and 0.28 seconds/image respectively. Matching one probe image with one gallery image is very efficient because it is performed as a dot product of a feature vector and regression coefficients (Section \ref{wristMatching}), which takes 0.0004 seconds/image. In the experiment with the largest number of wrists in the gallery set (EXP3), the average time of Meta-recognition stage (Section \ref{metaRecognition}) is 0.005 seconds/image. The computational complexity of the WMM is linear with respect to the number of gallery wrists because it requires $n$ comparisons of probe and gallery wrists, where $n$ is a number of gallery wrists.

\begin{table}[]

\renewcommand{\arraystretch}{1.1}
\caption{Details of three different experiments.}
\label{tabExpScenarios}
\centering
\begin{tabular}{c|c|c|c}
\hline
\bfseries Exp & \bfseries Gallery &\bfseries Probe &\bfseries Forensic cases related to:\\
\hline\hline
EXP1 & SET1 & SET2 & Child pornography. Good image quality, \\ &&& no much pose or  viewpoint variation. \\
\hline
EXP2 & SET1 & SET3 & Child pornography, rioter images. \\ &&&  With sufficient resolution but with\\ &&&  pose, viewpoint, illumination variations. \\
\hline
EXP3 & SET4 & SET5 & Terrorist images. Low resolution, pose, \\ &&& viewpoint, illumination, shadowing variations. \\
\hline
\end{tabular}
\end{table}

\subsection{Evaluation of the WMFA and WMM}\label{WMeva}
\begin{figure*}[]
\centering
\subfloat[]{\includegraphics[width=3in]{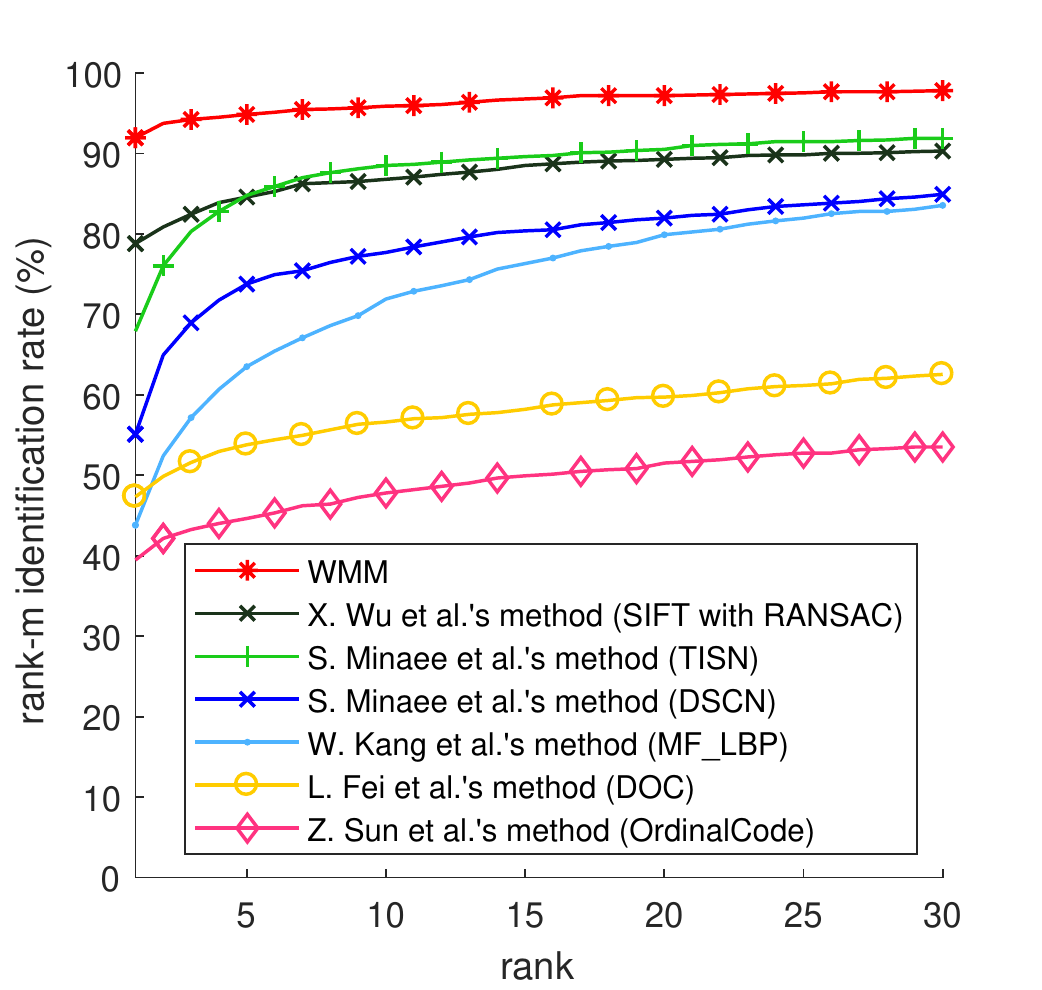}
\label{fig1exp1}}
\hfil
\subfloat[]{\includegraphics[width=3in]{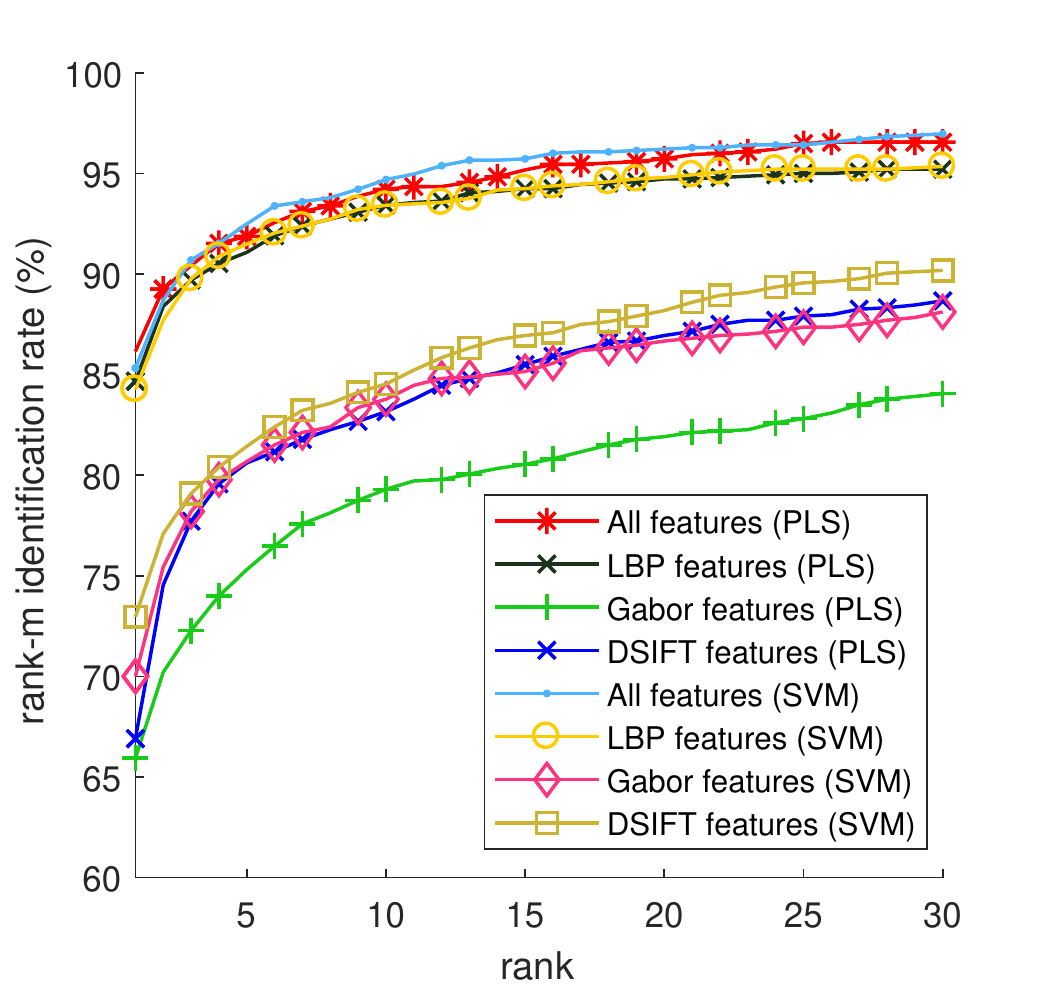}
\label{fig2exp1}}
\hfil
\subfloat[]{\includegraphics[width=3in]{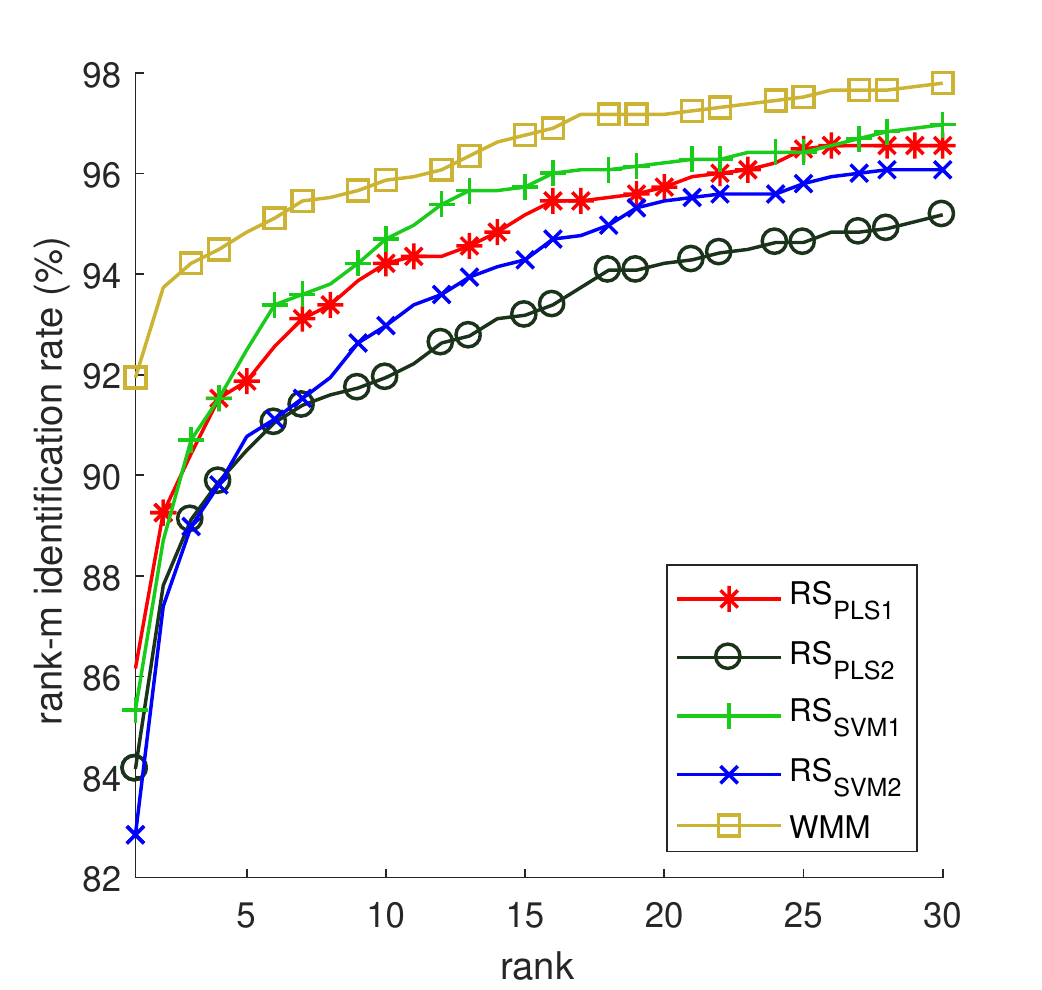}
\label{fig3exp1}}
\hfil
\subfloat[]{\includegraphics[width=3in]{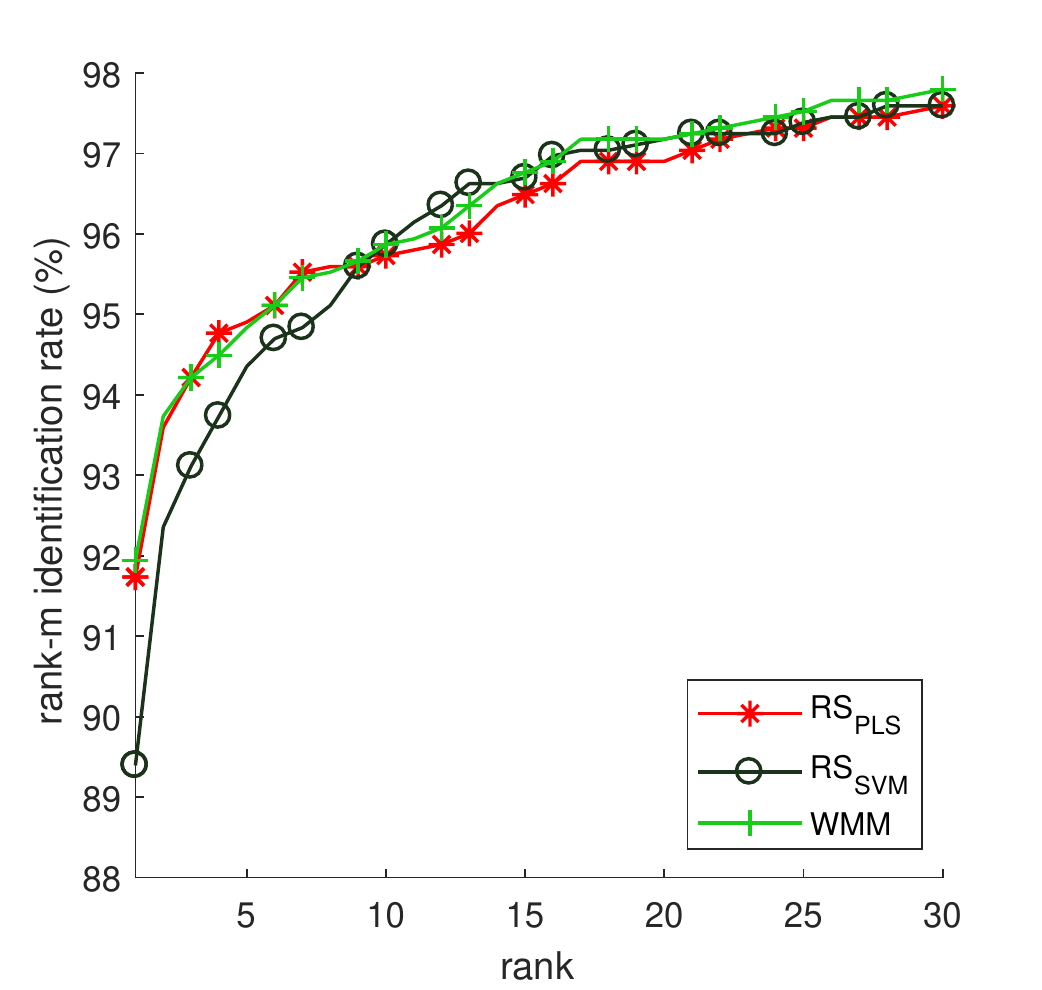}
\label{fig4exp1}}
\caption{CMC curves for EXP1 of (a) the WMFA algorithm and six different palmprint and palm vein matching methods, (b) particular feature using $RS_{PLS1}$ and $RS_{SVM1}$, (c) different recognition systems and (d) Meta-recognition systems.}
\label{figExp1}
\end{figure*}

\begin{figure*}[]
\centering
\subfloat[]{\includegraphics[width=3in]{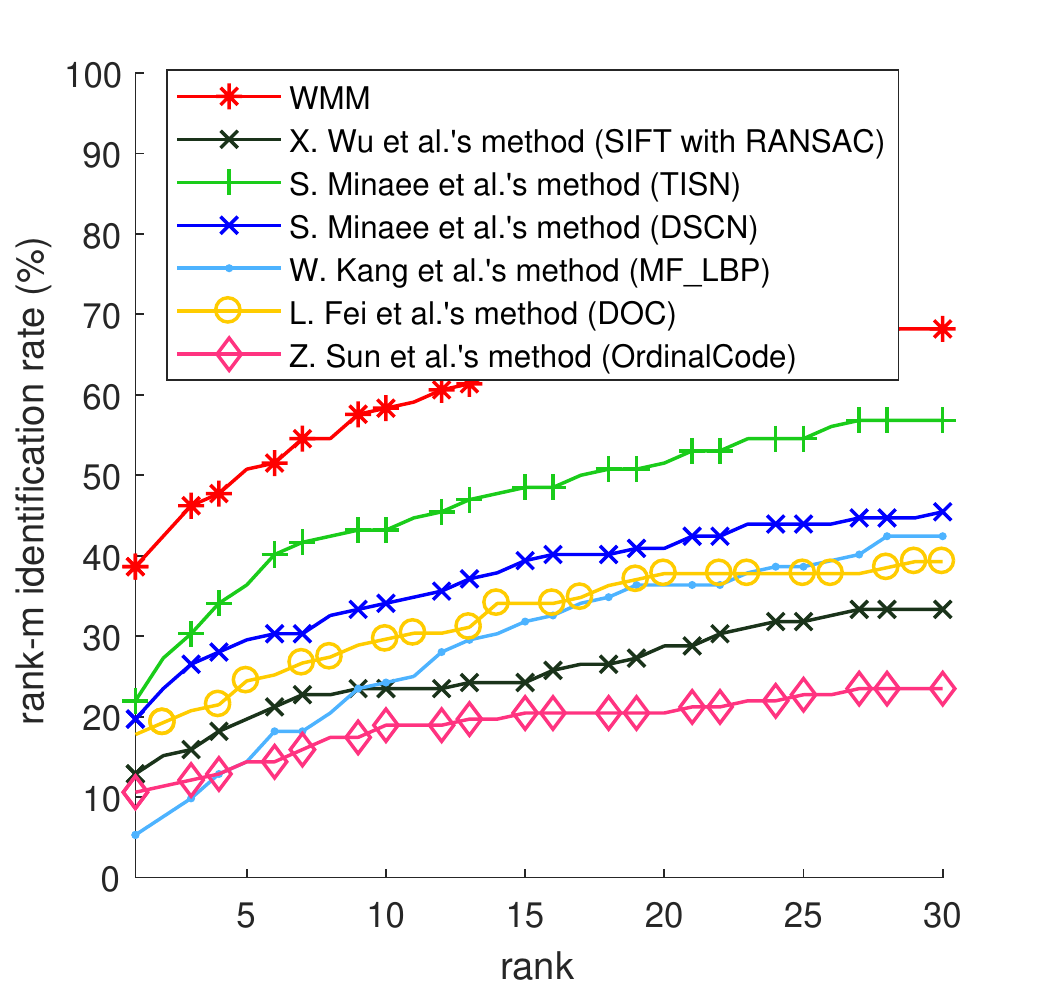}
\label{fig1exp2}}
\hfil
\subfloat[]{\includegraphics[width=3in]{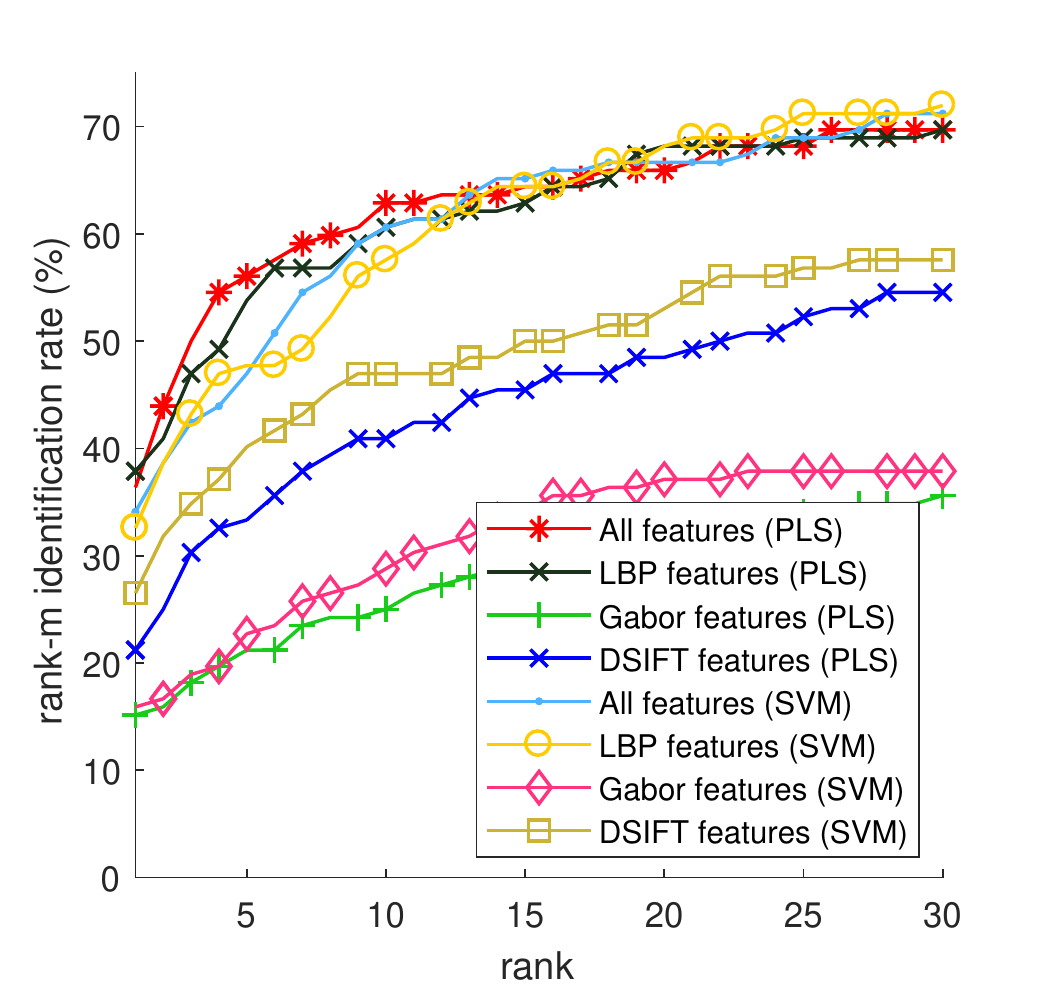}
\label{fig2exp2}}
\hfil
\subfloat[]{\includegraphics[width=3in]{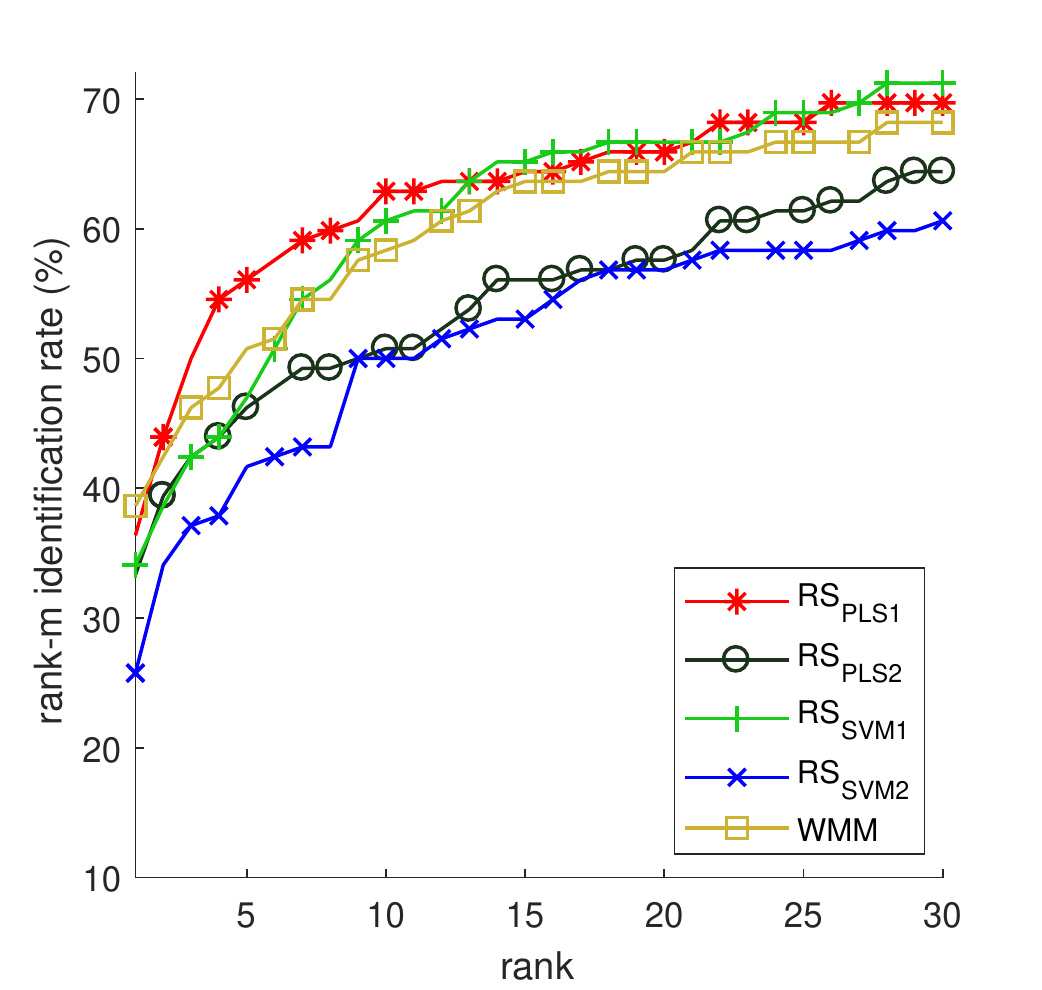}
\label{fig3exp2}}
\hfil
\subfloat[]{\includegraphics[width=3in]{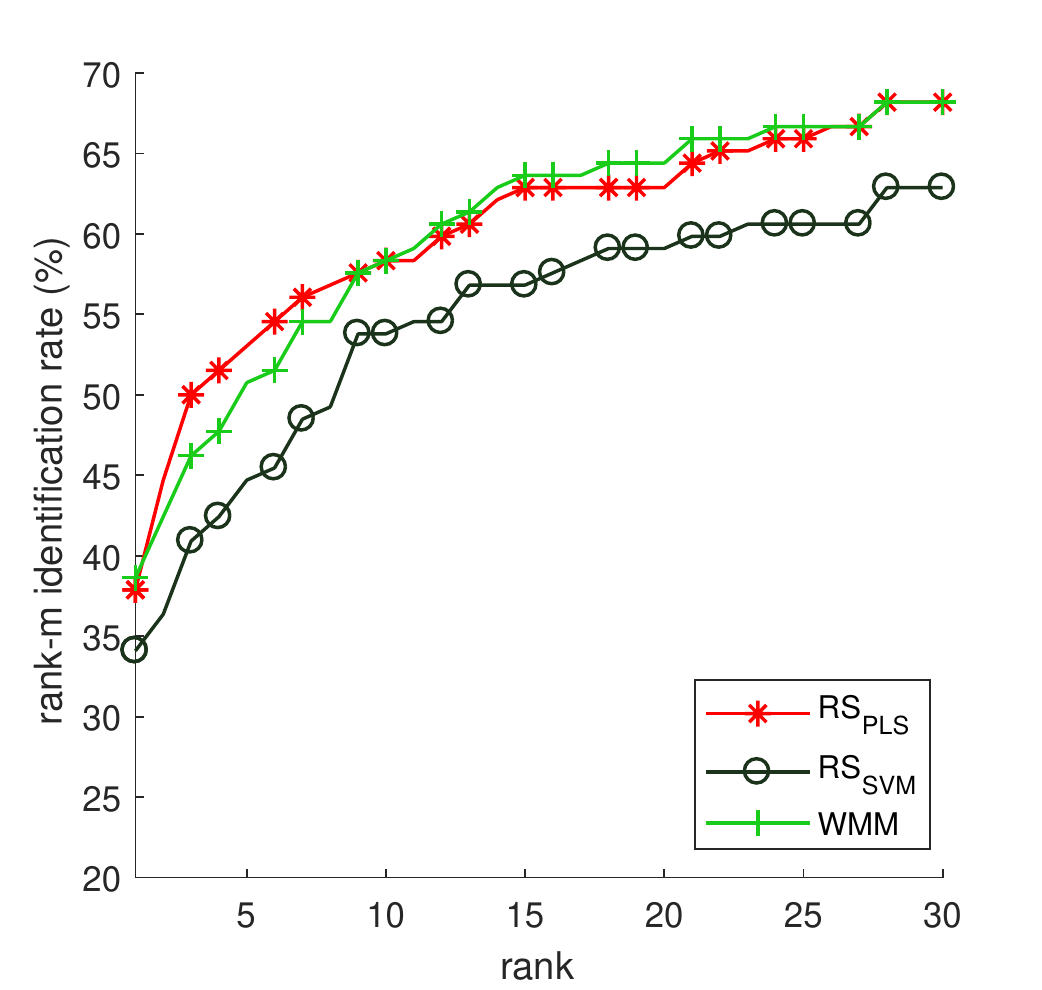}
\label{fig4exp2}}
\caption{CMC curves for EXP2 of (a) the WMFA algorithm and six different palmprint and palm vein matching methods, (b) particular feature using $RS_{PLS1}$ and $RS_{SVM1}$, (c) different recognition systems and (d) Meta-recognition systems.}
\label{figExp2}
\end{figure*}

\begin{figure*}[]
\centering
\subfloat[]{\includegraphics[width=3in]{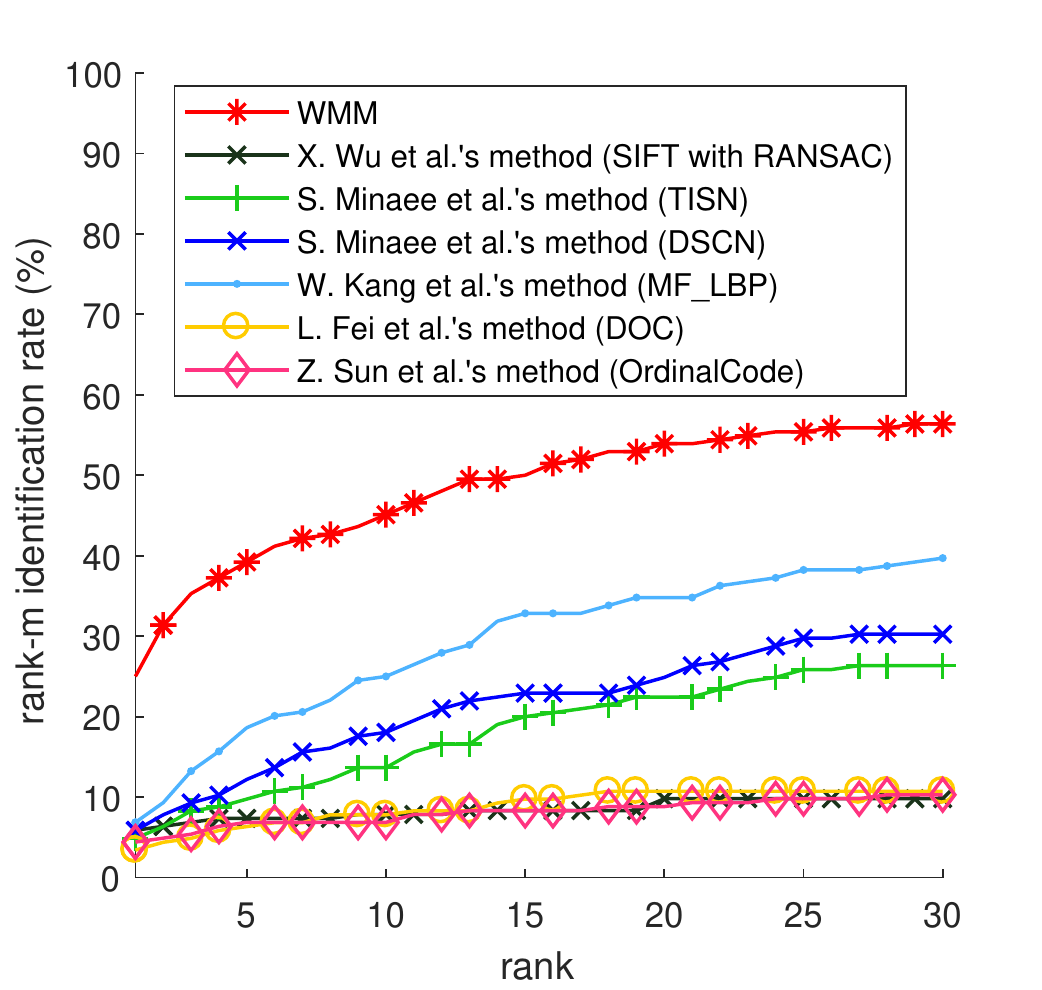}
\label{fig1exp3}}
\hfil
\subfloat[]{\includegraphics[width=3in]{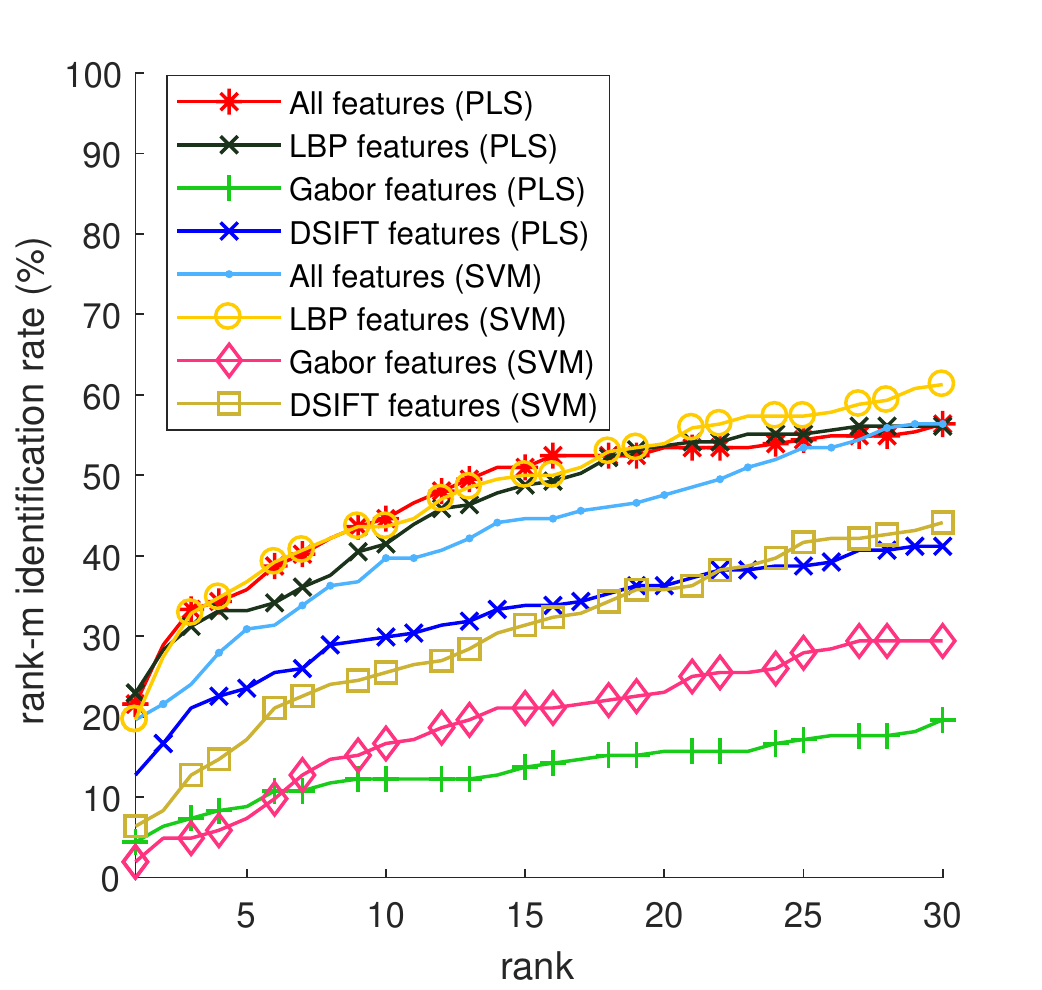}
\label{fig2exp3}}
\hfil
\subfloat[]{\includegraphics[width=3in]{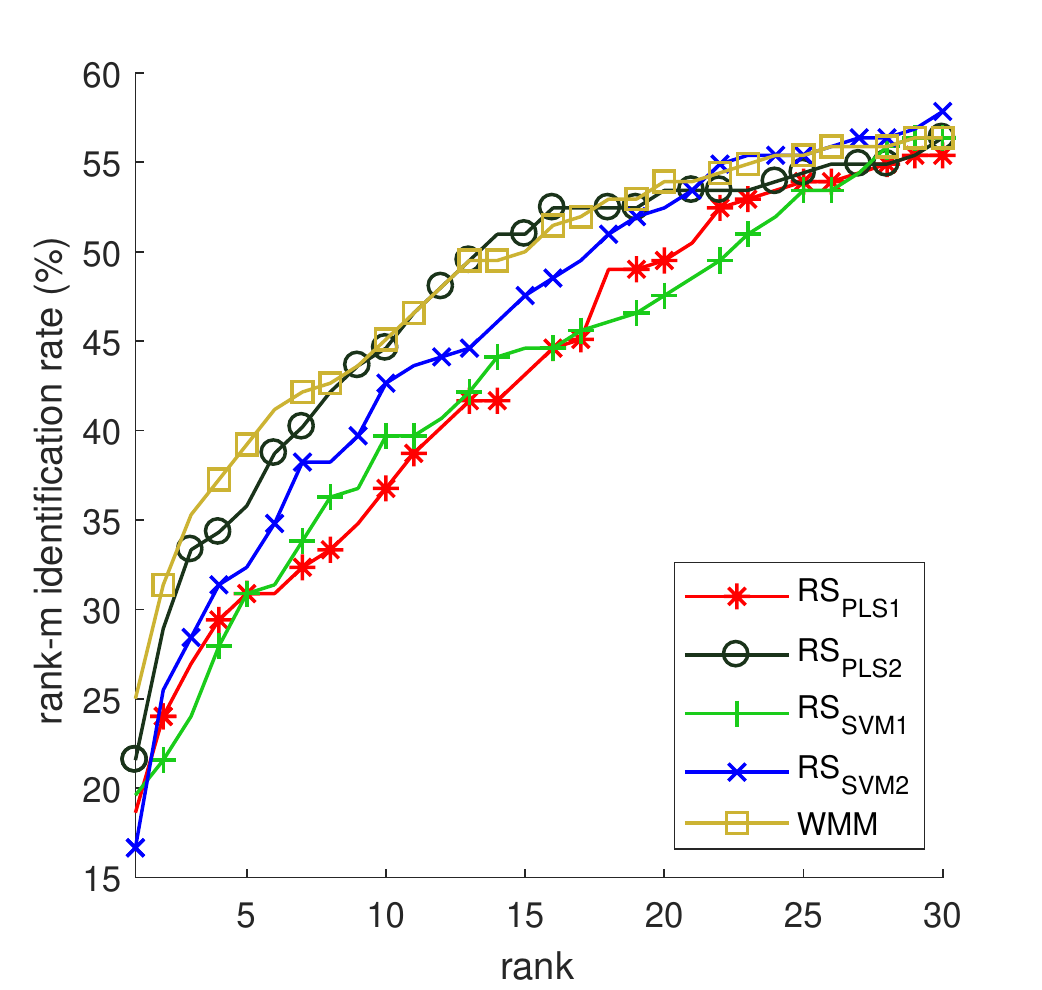}
\label{fig3exp3}}
\hfil
\subfloat[]{\includegraphics[width=3in]{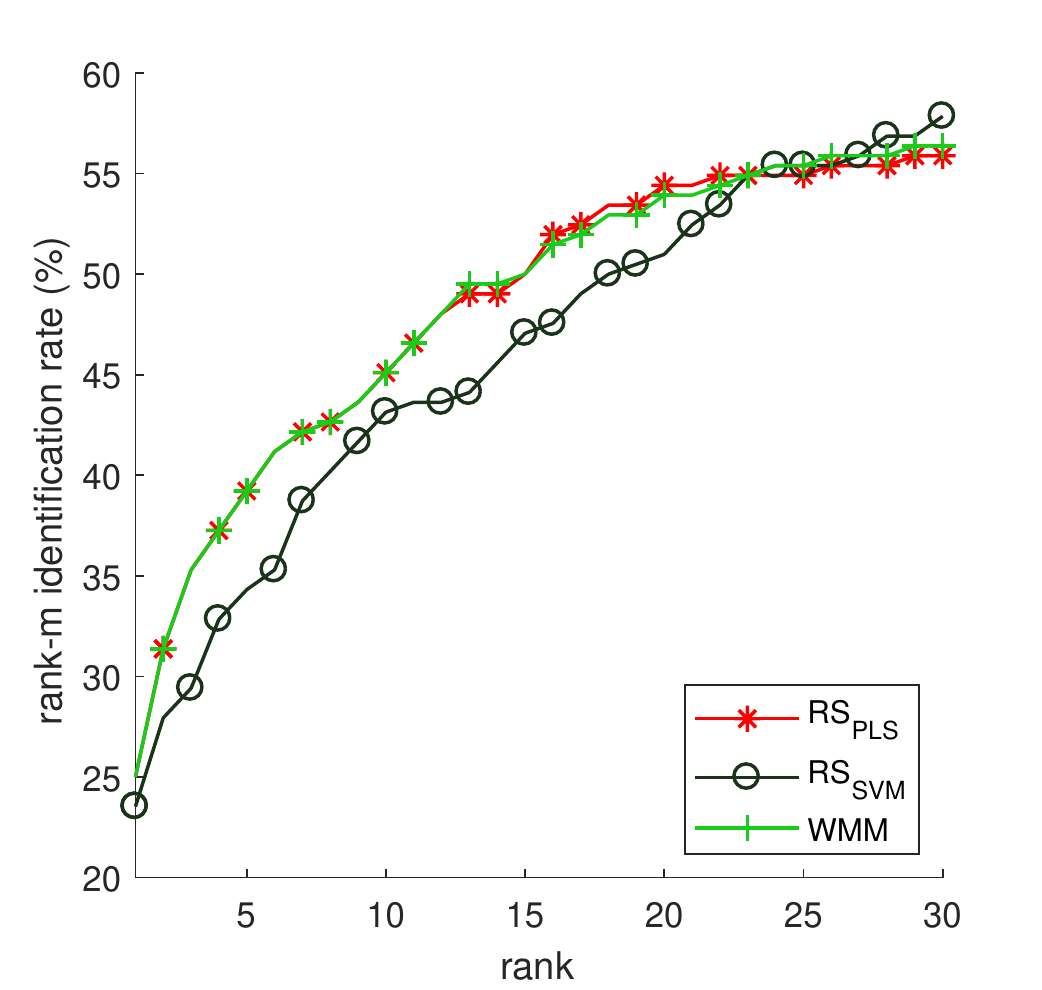}
\label{fig4exp3}}
\caption{CMC curves for EXP3 of (a) the WMFA algorithm and six different palmprint and palm vein matching methods, (b) particular feature using $RS_{PLS2}$ and $RS_{SVM1}$, (c) different recognition systems and (d) Meta-recognition systems.}
\label{figExp3}
\end{figure*}

\begin{figure}[]
\centering
\subfloat[]{\includegraphics[width=6in]{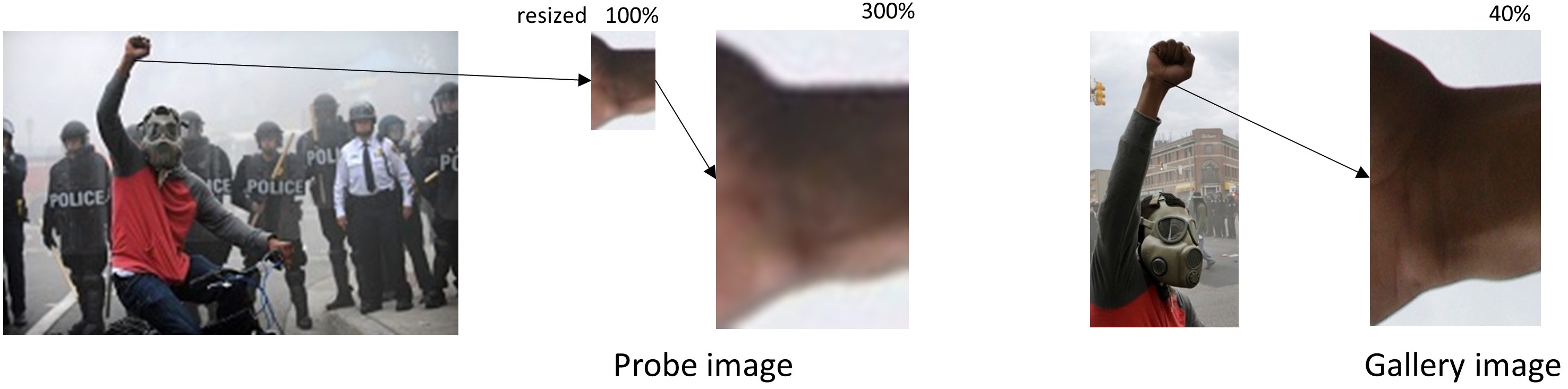}
\label{figProbeBaltimore}}

\subfloat[]{\includegraphics[width=6in]{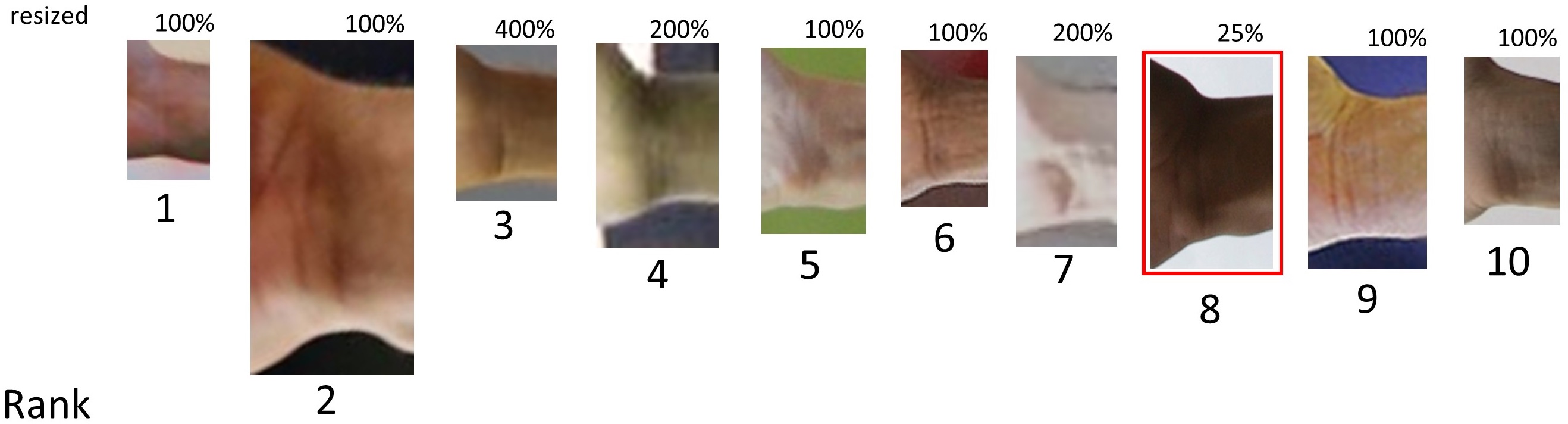}
\label{figRank8Baltimore}}

\subfloat[]{\includegraphics[width=6in]{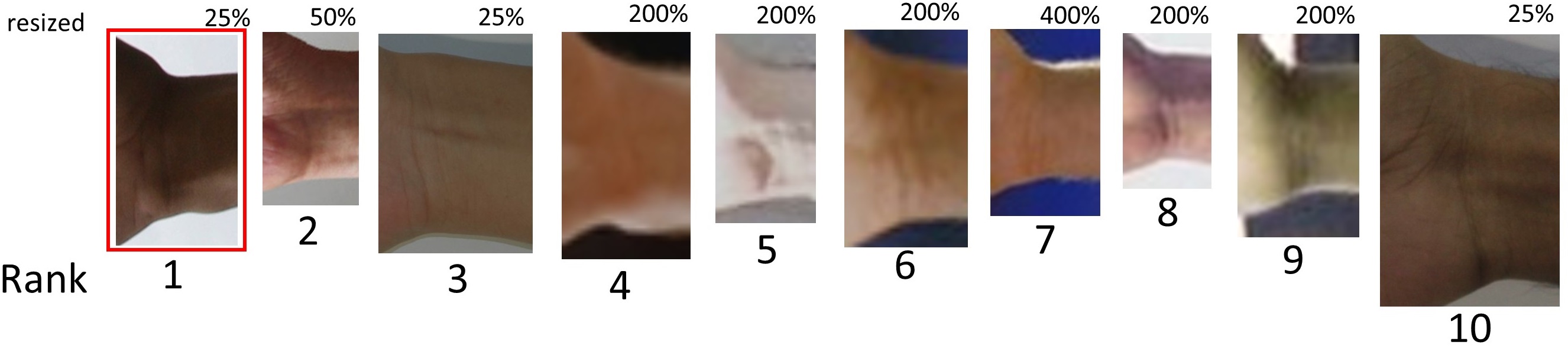}
\label{figRank1Baltimore}}
\caption{Top 10 matches returned by the WMM algorithm in EXP3 when the tail size is 50\% of the gallery size (b) and 10\% (c). The genuine match is highlighted with a red box. Probe and gallery Internet images taken outdoor containing the Baltimore rioter are shown in (a). Wrist images are resized for better visualization. The scale is indicated above each image.}
\label{baltimoreRank}
\end{figure}

In the first experiment EXP1, both sets are standard pose images. Gallery SET1 is used to build 526 classifiers from 1948 images and probe SET2 contains 1452 images for testing. The gallery set is larger than the probe set in terms of classes. It has 526 different classes while the probe set has 397 corresponding classes. 
The experimental results in Fig. \ref{fig1exp1} show that the WMM and WMFA algorithms outperform the state-of-the-art palmprint and palm vein matching methods by achieving accuracy of 91.94\% at rank-1 and 97.80\% at rank-30. Feature contribution is evaluated for two different WMFAs, $RS_{PLS1}$ and $RS_{SVM1}$ because they achieve better result than $RS_{PLS2}$ and $RS_{SVM2}$. The most discriminative features are LBP, then SVM's DSIFT and Gabor (Fig. \ref{fig2exp1}). The four WMFAs $RS_{PLS1}$, $RS_{PLS2}$, $RS_{SVM1}$ and $RS_{SVM2}$ are compared to evaluate PLS and SVM classifiers and the ROI extraction procedures. Fig. \ref{fig3exp1} shows that WMFAs with SVMs generally perform worse than their corresponding WMFAs with PLS and the highest rank-1 accuracy of 86.16\% and 85.33\% is achieved by respectively $RS_{PLS1}$ and $RS_{SVM1}$ indicating that ROI\#1 is more robust than ROI\#2. Comparing with the four WMFAs, the proposed WMM ($RS_{PLS+SVM}$) provides 5.78\% improvement at rank-1, which demonstrates the effectiveness of the Meta-recognition scheme based on EVT. Fig.  \ref{fig4exp1} compares WMM, $RS_{PLS}$ and $RS_{SVM}$, all based on the Meta-recognition scheme, showing that $RS_{PLS}$ outperforms $RS_{SVM}$ with a margin of  2.35\% at rank-1. \\
\indent
In experiment EXP2 the gallery set remains the same as in EXP1, but 135 images from SET3 are used as a probe set. The probe set contains 133 unique wrists which means that only 2 wrists have 2 images and the rest have one image per wrist. EXP2 is more challenging than EXP1 because the probe images usually have variation due to illumination, pose or out of focus. Some of the example images can be found in Fig. \ref{fig2dbLowQualitySet3}. As shown in  Fig. \ref{fig1exp2}, the proposed WMM algorithm also significantly outperforms the state-of-the-art palmprint and palm vein matching methods with a margin of 16.67\% in terms of rank-1 accuracy. It achieves rank-1 accuracy of 38.64\% and rank-30 accuracy of 68.18\%. Similarly to EXP1, the most discriminative features are LBP and ROI\#1 as shown in Fig. \ref{fig2exp2}. Comparison of WMFAs, including $RS_{PLS1}$, $RS_{PLS2}$, $RS_{SVM1}$ and $RS_{SVM2}$, is presented in Fig. \ref{fig3exp2} and similarly to EXP1 it shows that ROI\#1 is more robust than ROI\#2. Fig. \ref{fig4exp2} shows that WMM algorithm, which considers all $RS$, achieves the highest accuracies for rank-1: 38.64\%. Unlike in EXP1, at rank-15, WMM performs 0.75\%, 1.51\% worse than $RS_{PLS1}$  and $RS_{SVM1}$ respectively which is presented in Table \ref{tabResults}. The Meta-recognition scheme considers the top matching score and therefore, it can provide improvement at rank-1. However, there is no theoretical guarantee that it would improve accuracy of other ranks. More discussion about the performance of the Meta-recognition step is given in Section \ref{discussion}.\\
\indent
In experiment EXP3, the standard pose gallery set is enlarged by adding 205 Internet images to create gallery SET4, which contains 2153 images from 731 different wrists. 205 probe images in SET5 are only Internet images and each of them is from a corresponding wrist in the gallery. This experiment is the most challenging one because the Internet images were taken under extremely uncontrolled environments. Their variations come from uneven illumination, shading, pose, out of focus, point of view, and low resolution. Some examples are given in Figs. \ref{fig3dbInternetSet4} and \ref{fig3dbInternetSet5}. Moreover, each of the Internet image classifiers was trained based on one positive sample and 2152 negative samples coming from the standard pose and Internet images. PLS is capable to handle this imbalanced classification problem \cite{Qu2010}. As mentioned before, the standard pose images have more than one positive sample image to train the classifiers. Fig. \ref{fig1exp3} shows that the WMM algorithm achieves rank-1 accuracy of 24.88\% and rank-15 accuracy of 50.73\%, which significantly outperforms the other methods with a margin of  18.02\% at rank-1 accuracy. For the best WMFAs - $RS_{PLS2}$ and $RS_{SVM1}$, feature channels are evaluated and similarly to EXP1 and EXP2, the most discriminative features are LBP and Gabor features perform the worst (Fig. \ref{fig2exp3}). Comparison of different WMFAs shows again that PLS  achieve better results (Fig. \ref{fig3exp3}). Fig. \ref{fig4exp3} shows that Meta-recognition step increases rank-1 accuracies for both $RS_{PLS}$ and $RS_{SVM}$ but neither improves nor degrades WMM, comparing with $RS_{PLS}$ at most of the ranks. WMM performs the same as $RS_{PLS}$ until rank-12, though WMM performs 0.5\% better at rank-30. \\
\indent
Some of the top 10 matches are presented in Figs. \ref{baltimoreRank} and \ref{top10Ranks}. The proposed algorithm is able to retrieve the masked Baltimore rioter within top 10 ranks (see Fig. \ref{baltimoreRank}). The resolution of the masked Baltimore rioter in the gallery set is 198 by 342 pixels, while the resolution of the corresponding probe image is only 19 by 45 pixels (see Fig. \ref{figProbeBaltimore}). As shown in Figs. \ref{figRank8Baltimore} and \ref{figRank1Baltimore}, WMM matches them at rank-8 and rank-1 when 50\% and 10\% of the tail of the matching scores are respectively used to estimate the parameters of the Weibull distribution. However decreasing the tail size from 50\% to 10\%  slightly decreases general performance of the WMM achieving accuracy of 24.39\%, 46.83\% and 53.17\% at respectively rank-1, rank-15 and rank-30. $RS_{PLS1}$, $RS_{PLS2}$, $RS_{SVM1}$ and $RS_{SVM2}$ rank Baltimore rioter image at respectively the 4\textsuperscript{th}, 8\textsuperscript{th}, 1\textsuperscript{st}, and 3\textsuperscript{rd} positions. This example exposes the potential of wrists for criminal and victim identification.

\begin{table}[]
\renewcommand{\arraystretch}{1.1}
\caption{Experimental results - methods comparison.}
\footnotesize
\label{tabResults}
\centering
\begin{tabular}{c|c|c|c|c|c|c|c|c|c|}
\cline{2-10}
& \multicolumn{3}{ c|}{\bfseries EXP1} & \multicolumn{3}{ c| }{\bfseries EXP2} & \multicolumn{3}{ c| }{ \bfseries EXP3}\\
\cline{1-10}
\bfseries \scalebox{1.2}{Method} & rank 1 & rank 15 & rank 30 & rank 1 & rank 15 & rank 30 & rank 1 & rank 15 & rank 30 \\
\hline 
\hline
WMM & \underline{\textbf{91.94}} & \underline{\textbf{96.97}} & \underline{\textbf{97.80}} & \underline{\textbf{38.64}} & 63.64 & 68.18 & \underline{\textbf{24.88}} & 50.73 & 56.59 \\
\hline
\scalebox{.8}{$RS_{PLS}$} & 91.74 & 96.49 & 97.59 & 37.88 & 63.64 & 68.18 & \underline{\textbf{24.88}} & 50.73 & 56.10 \\
\hline
\scalebox{.8}{$RS_{SVM}$} & 89.39 & 96.69 & 97.59 & 32.58 & 56.82 & 62.88 & 23.90 & 46.83 & 57.58 \\
\hline
\scalebox{.8}{$RS_{PLS1}$} & 86.16 & 95.18 & 96.56 & 36.36 & 64.39 & 69.70 & 18.54 & 43.90 & 55.61 \\
\hline
\scalebox{.8}{$RS_{PLS2}$} & 84.16 & 93.18 & 95.18 & 33.33 & 56.06 & 64.39 & 22.93 & \underline{\textbf{51.71}} & 55.61 \\
\hline
\scalebox{.8}{$RS_{SVM1}$} & 85.33	& 95.73	& 96.97	& 34.09	& \underline{\textbf{65.15}}	& \underline{\textbf{71.21}}	& 20.00	& 44.88	& 56.10 \\
\hline
\scalebox{.8}{$RS_{SVM2}$} & 82.85	& 94.28	& 96.07	& 25.76	& 53.03	& 60.61	& 16.59	& 47.32	& \underline{\textbf{58.05}} \\
\hline
{\cite{Wu2014}}& 78.79 & 88.50 & 90.29 & 12.88 & 24.24 & 33.33 & 5.88 &	8.33 & 9.80\\
\hline
{\cite{Minaee2015}}& 67.91 & 89.60 & 91.87 & 21.97 & 48.48 & 56.82 & 4.39 & 18.05 & 28.78 \\
\hline 
{\cite{Minaee2017}}& 62.40 & 84.50 & 87.67 & 19.70 & 39.39 & 45.45 & 2.44 & 20.00 & 29.76 \\
\hline
{\cite{Fei2016}} & 47.31 & 58.20 & 62.53 & 17.78 & 34.07 & 39.26 & 3.42 & 9.76 & 10.73 \\
\hline
{\cite{Kang2014}}& 43.80 & 76.31 & 83.54 & 5.30 & 31.82 &	42.42 &	6.86 & 32.84 & 39.71\\
\hline
{\cite{ZhenanSun2005}}& 39.46 & 49.93 & 53.51 & 10.61 & 20.45 &	23.48 & 4.41 & 8.33 & 10.29 \\
\hline
\multicolumn{10}{c} \raggedleft{The highest percentage accuracy is highlighted.} 
\end{tabular}
\end{table}

\subsection{Dimension Reduction and Feature Selection}\label{PCA}
This subsection aims to compare the high-dimensional (16466-dimensional) and lower dimensional feature vectors  extracted from ROI\#1 and ROI\#2, which are used to train PLS and SVM classifiers. However, it should be pointed out that PLS regression performs a supervised dimension reduction internally \cite{Rosipal2006}. Thus it can be also considered as an internal dimension reduction scheme, controlled by the number of the latent components. Additional experiments are performed to investigate the effect of lower dimensional (200-dimensional) features. For dimension reduction, PCA with 200 components in four different $RS$ settings denoted as $PCA+PLS1$, $PCA+PLS2$, $PCA+SVM1$ and $PCA+SVM2$ is applied. For feature selection, minimal-redundancy-maximal-relevance (mRMR) criterion \cite{Peng2005} is used to select 200 features for $mRMR+PLS1$, $mRMR+PLS2$, $mRMR+SVM1$ and $mRMR+SVM2$. In Table \ref{tabResultsPCA}, the results  are reported as a difference between $PCA+PLS1$, $PCA+PLS2$, $PCA+SVM1$, $PCA+SVM2$ and the corresponding $RS_{PLS1}$, $RS_{PLS2}$, $RS_{SVM1}$ $RS_{SVM2}$, respectively. In Table \ref{tabResultsmRMR}, the results are presented in the same way as in Table \ref{tabResultsPCA}, but for the mRMR feature selection scheme. The negative and positive numbers indicate the performance drop and gain respectively. For example, -4.14 for $PCA+PLS1$ means that after applying PCA on the high-dimensional feature vector extracted from ROI\#1 and using PLS, the performance drops by 4.14\% with respect to $RS_{PLS1}$. Table \ref{tabResultsPCA} shows that applying PCA with 200 components mostly decreases the identification performance, especially for PLS in EXP1 and EXP2 and SVM in EXP3. In our experiments, similar performance degradation is also observed for a different number of principal components ranging from 100 to 1000 and thus the results with 200 components are provided. Table \ref{tabResultsmRMR} shows that the performance drop is even more significant, when mRMR is used to select 200 features. In \cite{Peng2005}, the authors used up to 50 selected features, even for 9703-dimensional feature vectors. However in our experiments with such low-dimensional feature vectors selected by mRMR, the performance degradation in EXP1, EXP2 and EXP3 is even higher. Thus 200 features are used for the fair comparison with PCA. Note that we do not claim that the high-dimensional features used in the WMFAs are the best wrist representation and other representations can be also considered. Nevertheless, in this study, the main focus is to show that wrist can be useful for forensic investigation. 

\begin{table}[]
\renewcommand{\arraystretch}{1.1}
\caption{The accuracy (\%) differences between PCA compressed features and the original high-dimensional features.}
\footnotesize
\label{tabResultsPCA}
\centering
\begin{tabular}{c|c|c|c|c|c|c|c|c|c|}
\cline{2-10}
& \multicolumn{3}{ c|}{\bfseries EXP1} & \multicolumn{3}{ c| }{\bfseries EXP2} & \multicolumn{3}{ c| }{ \bfseries EXP3}\\
\cline{1-10}
\bfseries \scalebox{1.2}{Method} & rank 1 & rank 15 & rank 30 & rank 1 & rank 15 & rank 30 & rank 1 & rank 15 & rank 30 \\
\hline 
\hline
\scalebox{.8}{$PCA+PLS1$} & -4.14 & -1.04 & -0.77 & -5.30 & -3.03 & -3.80 & -1.47 & -1.47 & -1.47 \\
\hline
\scalebox{.8}{$PCA+PLS2$} & -6.75 & -3.24 & -2.69 & -12.88 & -6.82 & -2.76 & -4.40 & -2.93 & 2.92 \\
\hline
\scalebox{.8}{$PCA+SVM1$} & -0.97	& -1.45	& -0.83	& 2.27	& -7.58	& -6.82	& -9.27	& -11.71 & -12.20 \\
\hline
\scalebox{.8}{$PCA+SVM2$} & -0.42	& -1.45	& -2.07	& -0.01	& -0.76	& -3.80	& -5.42	& -7.32	& -9.76 \\
\hline
\end{tabular}
\end{table}

\begin{table}[]
\renewcommand{\arraystretch}{1.1}
\caption{The accuracy (\%) differences between features selected by mRMR and the original high-dimensional features.}
\footnotesize
\label{tabResultsmRMR}
\centering
\begin{tabular}{c|c|c|c|c|c|c|c|c|c|}
\cline{2-10}
& \multicolumn{3}{ c|}{\bfseries EXP1} & \multicolumn{3}{ c| }{\bfseries EXP2} & \multicolumn{3}{ c| }{ \bfseries EXP3}\\
\cline{1-10}
\bfseries \scalebox{1.2}{Method} & rank 1 & rank 15 & rank 30 & rank 1 & rank 15 & rank 30 & rank 1 & rank 15 & rank 30 \\
\hline 
\hline
\scalebox{.8}{$mRMR+PLS1$} & -22.52 & -8.47 & -5.24 & -22.72 & -23.48 & -18.18 & -7.81 & -8.30 & -10.74 \\
\hline
\scalebox{.8}{$mRMR+PLS2$} & -26.52 & -8.47 & -5.24 & -22.72 & -23.48 & -18.18 & -7.81 & -8.30 & -10.74 \\
\hline
\scalebox{.8}{$mRMR+SVM1$} & -14.46	& -8.06	& -4.75	& -19.70	& -19.70	& -18.94	& -11.22	& -7.32 & -9.76 \\
\hline
\scalebox{.8}{$mRMR+SVM2$} & -18.32	& -9.36	& -6.61	& -8.34	& -17.42	& -20.46	& -6.83	& -12.20	& -8.78 \\
\hline
\end{tabular}
\end{table}

\subsection{Deep vs Hand Crafted Features}\label{expFeat}
This set of experiments aims to evaluate the deep features (DF) from \cite{Minaee2015} and \cite{Minaee2017} denoted as df1 and df2 respectively, mdLBP from \cite{Dubey2014} and their fusion with the hand crafted features employed in WMFAs and WMM denoted as HC. The HC consists of LBP, Gabor and DSIFT described in Section \ref{featureExtraction}.
The fusion is indicated by "+" sign. For example, df1+ and mdLBP+ mean that the feature vector consists of df1 with HC and mdLBP with HC respectively. The same set of experiments, EXP1, EXP2 and EXP3 with different settings on the features is performed. Note that these experiments do not aim to compare different ROIs. Thus as mentioned before, for each feature and experiment, the results for the ROI with the highest rank-1 accuracy are given. In addition, the results of two representative deep learning methods AlexNet \cite{Krizhevsky2012} and VGG-16 \cite{Simonyan2015} are also reported in Table \ref{tabResultsDL}.\\
\indent
The results of LBP, mdLBP and mdLBP+ are reported in Table \ref{tabResultsFeatures}a. Employing mdLBP (34864-dimensional) or mdLBP+ (51330-dimensional) generally neither improves the performance nor outperforms LBP (13074-dimensional) and HC (16466-dimensional). Note that similarly to LBP, mdLBP features are also extracted within each block using the seven grids shown in Fig. \ref{figGrids}.
There are three cases when mdLBP+ with SVM outperforms LBP or HC.
In EXP1, for rank-15 and rank-30 mdLBP+ performs 0.55\% and 0.62\% higher than LBP and for rank-30 mdLBP+ achieves 0.34\% higher accuracy than HC ($RS_{SVM1}$).\\
\indent
The deep features df1 (782-dimensional) and df2 (12512-dimensional) are extracted from the first and the second layer of the scattering network which is a DCN that instead of learning the filters uses predefined wavelets \cite{Bruna2013}. In \cite{Minaee2015} the mean and variance over the whole scattering transformed images are calculated to form df1, whereas in \cite{Minaee2017} the same statistics are calculated from 16 blocks (4x4 grid) over the transformed images to form df2. In these experiments, no PCA is used because PCA compressed features with 200 principal components from \cite{Minaee2015} and \cite{Minaee2017}, perform worse than df1 and df2, respectively (see Table \ref{tabResults} and \ref{tabResultsFeatures}b). Note that the features from \cite{Minaee2015} and \cite{Minaee2017} without PCA are equivalent to df1 and df2, respectively.
The comparison of DF, HC and the fusion (df1+, df2+) using SVM and PLS is presented in Table \ref{tabResultsFeatures}b. In all the experiments HC outperforms DF. Given the same type of classifier, df2 is superior to df1. Employing DF with SVM always gives better results than PLS. The feature fusions of DF with HC: df1+ (17248-dimensional) and df2+ (28978-dimensional) always outperform df1 and df2. Generally, df1+'s performance is higher than df2+ but for rank-1s, HC still achieves higher accuracies than df1+. In EXP2 and EXP3, for rank-15s and rank-30s, df1+ achieves around 1\% and 3\% higher accuracy than HC respectively. 
As shown in Table \ref{tabResultsFeatures}c if there is the Meta-recognition step on top of the systems, HC outperforms df1+ and df2+, except for EXP1 rank-15 and EXP3 rank-30, where they perform the same. \\
\indent
It should be emphasized that in the deep scattering network, the filters are not learnt but fixed. Thus, in the experiments, there is no learning for df1 and df2. Training deep learning architectures requires sufficient number of training examples, whereas the number of wrist examples in NTU-Wrist-Image-Database-v1 is not very large (see Table \ref{tabDB2}). Mostly, there are four images per wrist in the standard pose SET1 and SET2 whereas the Internet images in SET1P and SET5 have only one image per wrist. Nevertheless, two well-known deep learning architectures AlexNet \cite{Krizhevsky2012} and VGG-16 \cite{Simonyan2015} are selected to investigate whether the training datasets in NTU-Wrist-Image-Database-v1 are able to support these deep learning methods. The proposed segmentation and ROI extraction methods are used to pre-process the images before training and testing. Additionally, the extracted ROI (ROI\#1) images are padded with zeros to make them square because AlexNet and VGG-16 require fixed size, square images. First, the networks are pretrained on ImageNet dataset. Then, the networks are fine-tuned until convergance which typically occurcs after 25 epochs, with batch size 128, momentum 0.9, no weight decay and ADAM optimizer \cite{Kingma2014}. The learning rate is 0.001 for the last layer and 10 times smaller for other layers. To prevent overfitting, 0.5 dropout \cite{Srivastava2014} is applied before the last fully connected layer. The rank-1, rank-15 and rank-30 accuracy are presented in Table \ref{tabResultsDL}. The results show that all the proposed WMFAs and WMM perform significantly better than both deep learning methods. In addition, AlexNet and VGG-16 are outperformed by almost all other methods used for comparison (see Tables \ref{tabResultsDL} and \ref{tabResults}). NTU-Wrist-Image-Database-v1 is relatively small comparing to other databases used for deep learning. Thus, it may not be able to support these deep learning methods which require larger number of training examples.     

\begin{table}[]
\renewcommand{\arraystretch}{1.1}
\caption{Experimental results - AlexNet and VGG-16.}
\footnotesize
\label{tabResultsDL}
\centering
\begin{tabular}{c|c|c|c|c|c|c|c|c|c|}
\cline{2-10}
& \multicolumn{3}{ c|}{\bfseries EXP1} & \multicolumn{3}{ c| }{\bfseries EXP2} & \multicolumn{3}{ c| }{ \bfseries EXP3}\\
\cline{1-10}
\bfseries \scalebox{1.2}{Method} & rank 1 & rank 15 & rank 30 & rank 1 & rank 15 & rank 30 & rank 1 & rank 15 & rank 30 \\
\hline 
\hline
\scalebox{.8}{AlexNet} & 23.21 & 58.06 & 69.15 & 2.27 & 14.39 & 25.00 & 5.58 & 24.39 & 32.68 \\
\hline
\scalebox{.8}{VGG-16} & 43.04 & 79.89 & 86.85 & 4.54 & 16.67 & 28.79 & 5.36 & 25.37 & 36.10 \\
\hline 
\end{tabular}
\end{table}

\begin{table}[]

\renewcommand{\arraystretch}{1.1}
\caption{Experimental results - features comparison.}
\footnotesize
\label{tabResultsFeatures}
\centering
\begin{tabular}{c|c|c|c|c|c|c|c|c|c|c|}

\multicolumn{7}{c}\raggedright{(a) LBP VS mdLBP}\\
\cline{3-11}
\multicolumn{2}{c|}{} & \multicolumn{3}{ c|}{\bfseries EXP1} & \multicolumn{3}{ c| }{\bfseries EXP2} & \multicolumn{3}{ c| }{\bfseries EXP3}\\
\cline{1-11}
\scalebox{.9}{\textbf{F}} & \scalebox{.7}{\textbf{C}} & rank 1 & rank 15 & rank 30 & rank 1 & rank 15 & rank 30 & rank 1 & rank 15 & rank 30 \\
\hline
\scalebox{.8}{LBP} & \scalebox{.8}{P} & \underline{\textbf{85.12}} & 94.63 & 96.14 & \underline{\textbf{37.88}} & 62.88 & 69.70 & \underline{\textbf{22.93}} & \underline{\textbf{48.78}} & \underline{\textbf{56.10}} \\
\hline
\scalebox{.8}{LBP} & \scalebox{.8}{S} & 83.82 & 95.11 & 96.69 & 32.58 & \underline{\textbf{64.39}} & \underline{\textbf{71.97}} & 20.00 & 43.41 & 54.14 \\
\hline
\scalebox{.7}{mdLBP} & \scalebox{.8}{P} & 70.94 & 91.25 & 94.01 & 25.00 & 52.27 & 62.12 & 13.66 & 37.07 & 47.80 \\
\hline
\scalebox{.7}{mdLBP} & \scalebox{.8}{S} & 77.20 & 94.56 & 96.49 & 27.27 & 62.88 & 68.94 & 12.20 & 37.07 & 49.27 \\
\hline
\scalebox{.63}{mdLBP+} & \scalebox{.8}{P} & 82.78 & 94.70 & 96.14 & 26.53 & 49.24 & 56.06 & 16.10 & \underline{\textbf{48.78}} & 54.63 \\
\hline 
\scalebox{.63}{mdLBP+} & \scalebox{.8}{S} & 83.47 & \underline{\textbf{95.66}} & \underline{\textbf{97.31}} & 26.53 & 53.03 & 60.61 & 15.61 & 41.96 & 54.63 \\
\hline 
\multicolumn{11}{c}{}\\
\multicolumn{9}{c}\raggedleft{(b) DEEP VS HAND CRAFTED FEATURES in WMFA} \\
\cline{3-11}
\multicolumn{2}{c|}{} & \multicolumn{3}{ c|}{\bfseries EXP1} & \multicolumn{3}{ c| }{\bfseries EXP2} & \multicolumn{3}{ c| }{\bfseries EXP3}\\
\cline{1-11}
\scalebox{.9}{\textbf{F}} & \scalebox{.7}{\textbf{C}} & rank 1 & rank 15 & rank 30 & rank 1 & rank 15 & rank 30 & rank 1 & rank 15 & rank 30 \\
\hline
\scalebox{.8}{df2}  & \scalebox{.8}{S} & 78.65 & 91.94 & 93.53 & 30.30 & 55.30 & 58.33 & 7.70 & 32.30 & 43.90 \\
\hline
\scalebox{.8}{df2}  & \scalebox{.8}{P} & 61.91 & 87.60 & 91.12 & 12.12 & 41.67 & 48.48 & 5.85 & 32.68 & 44.88 \\
\hline
\scalebox{.8}{df1}  & \scalebox{.8}{S} & 73.48 & 89.53 & 91.05 & 25.00 & 47.73 & 53.79 & 6.82 & 29.76 & 39.51 \\
\hline
\scalebox{.8}{df1}  & \scalebox{.8}{P} & 39.26 & 81.61 & 88.02 & 4.54 & 25.00 & 41.67 & 2.43 & 20.49 & 33.17 \\
\hline
\scalebox{.8}{df2+} & \scalebox{.8}{S} & 83.26 & 95.25 & 96.69 & 33.33 & 57.58 & 65.91 & 17.07 & 40.49 & 50.24 \\
\hline
\scalebox{.8}{df2+} & \scalebox{.8}{P} & 80.51 & 94.49 & 95.94 & 35.61 & 64.39 & 70.73 & 18.05 & 46.34 & 55.12 \\
\hline
\scalebox{.8}{df1+} & \scalebox{.8}{S} & 84.30 & 95.80 & 96.69 & 32.58 & \underline{\textbf{65.91}} & 71.97 & 20.00 & 44.88 & \underline{\textbf{59.02}} \\
\hline
\scalebox{.8}{df1+} & \scalebox{.8}{P} & 84.92 & 94.97 & 96.83 & 35.61 & 65.15 & \underline{\textbf{72.73}} & 20.98 & \underline{\textbf{52.68}} & 56.59 \\
\hline
\scalebox{.8}{HC} & \scalebox{.8}{S} & 85.33 & \underline{\textbf{95.73}} & \underline{\textbf{96.97}} & 34.09 & 65.15 & 71.21 & 20.00 & 44.88 & 56.10 \\
\hline
\scalebox{.8}{HC} & \scalebox{.8}{P} & \underline{\textbf{86.16}} & 95.18 & 96.56 & \underline{\textbf{36.36}} & 64.39 & 69.70 & \underline{\textbf{22.93}} & 51.57 & 55.61 \\
\hline
\multicolumn{11}{c}{}\\
\multicolumn{9}{c}\raggedleft{(c) DEEP VS HAND CRAFTED FEATURES in WMM} \\
\cline{3-11}
\multicolumn{2}{c|}{} & \multicolumn{3}{ c|}{\bfseries EXP1} & \multicolumn{3}{ c| }{\bfseries EXP2} & \multicolumn{3}{ c| }{\bfseries EXP3}\\
\cline{1-11}
\scalebox{.9}{\textbf{F}} & \scalebox{.7}{\textbf{C}} & rank 1 & rank 15 & rank 30 & rank 1 & rank 15 & rank 30 & rank 1 & rank 15 & rank 30 \\
\hline
\scalebox{.8}{df2+} &\scalebox{.64}{M} &  87.26 & 95.45 & 96.35 & 37.88 & 58.33 & 66.67 & 18.05 & 44.88 & \underline{\textbf{56.59}} \\
\hline
\scalebox{.8}{df1+} & \scalebox{.64}{M} & 91.18 & \underline{\textbf{96.97}} & 97.59 & 37.12 & 56.82 & 63.64 & 23.90 & 47.32 & 55.61 \\
\hline
\scalebox{.8}{WMM} & \scalebox{.64}{M} & \underline{\textbf{91.94}} & \underline{\textbf{96.97}} & \underline{\textbf{97.80}} & \underline{\textbf{38.64}} & \underline{\textbf{63.64}} & \underline{\textbf{68.18}} & \underline{\textbf{24.88}} & \underline{\textbf{50.73}} & \underline{\textbf{56.59}} \\
\hline
\multicolumn{11}{c}\raggedleft{The highest percentage accuracy is highlighted.}\\
\multicolumn{11}{c}\raggedleft{F - features, C - classifier, P - PLS, S - SVM, M - Meta-recognition.} 
\end{tabular}
\end{table}
\normalsize

\section{Discussion}\label{discussion}
Criminal and Victim identification is an important part of forensic investigation. Although digital images are a useful clue for law enforcement agencies, suspects can intentionally hide their faces and tattoos to prevent identification. To tackle this problem, skin marks, androgenic hair patterns and vein patterns are studied recently. However, they are not always observable e.g., when the suspects wear long sleeves. On the other hand, their wrists are still visible when the suspects are raising their hands, holding weapons, or posing some gestures. However, wrist recognition was neglected by the biometric and forensic community. According to our best knowledge, it is the first study on wrist identification for forensic application based on images. \\
\indent
For this study, the NTU-Wrist-Image-Database-v1 containing 3945 images from 731 wrists from 505 subjects is collected. Its size is comparable to other public biometrics databases e.g. CAISA Palmprint Image Database with 5502 images from 312 subjects, FVC2006 with 1800 images from 150 different fingerprints, UBRIS.v2 iris database with 522 different irises from 261 subjects. In forensic investigation, the suspect databases are not always very large because some other information can be used to reduce their size. For example, many countries have a list of terrorist suspects, who may be written supportive messages to terrorist groups online or friends or relatives of confirmed terrorists. In some cases, it is in fact one-to-one authentication \cite{Bouchrika2011}. \\
\indent
The proposed wrist identification algorithm is evaluated on the NTU-Wrist-Image-Database-v1. In the pre-processing stage, the segmentation scheme using an ensemble of decision trees provides reliable results. The wrinkles detected by the shortest paths in the graph and the boundaries are used to determine two ROIs. LBP, Gabor and DSIFT features are extracted to represent wrists as 16466-dimensional feature vectors. For each wrist, PLS and SVM classifiers are built using the one-against-all approach. In order to select the best ROI and matching methods and increase the accuracy of the whole system, Meta-recognition step is used on top of WMFAs. The experimental results show that the proposed WMM algorithm significantly outperforms the state-of-the-art palmprint and palm vein recognition methods, including Kang et al.'s Mutual Foreground (MF)-Based LBP with $\chi^2$ distance method \cite{Kang2014} and Wu et al.'s SIFT-based method \cite{Wu2014} for contactless palmprint recognition. In addition, WMM also performs better than Minaee et al.'s methods \cite{Minaee2015,Minaee2017} which employ deep architectures.
For wrist matching, PLS is compared with linear SVM and generally performs slightly better especially in lower ranks while SVM tends to achieve more accurate results in higher ranks. Kernel SVM, which projects features to a high dimensional space, is not employed because the dimension of the feature vectors is already high. LBP are the most discriminative features in the proposed algorithm; DSIFT features are the second one while the Gabor features are the last. PCA dimension reduction and feature selection based on mRMR criterion do not improve the performance of PLS and SVM classifiers, thus the high-dimensional feature vectors are used to keep the discriminatory power. Moreover, the experimental results show that LBP over each RGB channel performs better than inter-channel based descriptor such as mdLBP. The comparison of HC and DF from \cite{Minaee2015,Minaee2017} demonstrates that the proposed HC outperforms the DF. The Meta-recognition step based on EVT improves rank-1 accuracy. Though WMM uses four recognition systems, which increases the time complexity, there is no real time requirement and the accuracy is more important in forensic investigation. \\
\indent
One can say that the identification performance does not seem to be spectacular especially for the non-standard and Internet images. In forensic investigations, poor quality images are very common making the case very challenging and difficult to be analysed. Mature identification methods when applied to bad quality images usually achieve low accuracy. For example Paulino et al. applied commercial fingerprint matchers on latent fingerprints with ugly quality and achieved rank-1 accuracies of  21.2-34.1\% \cite{Paulino2013}. In forensic investigation, not only rank-1 is important but also a wider range of ranks e.g. rank-30 is also useful for an investigator to narrow down the suspect list instead of searching through the entire database. In the experiments, rank-1 accuracies for non-standard and Internet images are respectively 38.64\% and 24.88\% and rank-15 accuracies are over 60\% and 50\%, respectively. 

\section{Conclusion}\label{conclusion}
In this paper, the wrist identification for forensic is studied. A new wrist image database named NTU-Wrist-Image-Database-v1 is established. The wrist identification algorithm, including skin segmentation, key point localization, image to template alignment, matching schemes, and post-recognition score analysis is proposed. The experimental results show that the proposed WMM algorithm significantly outperforms representative state-of-the-art palmprint and palm vein recognition methods. The proposed WMM algorithm successfully matches the masked Baltimore rioter within top 10 ranks. In fact, he is retrieved at the 1st rank and the 8th rank depending on the algorithm parameter. It clearly shows that wrist is possible to be used for forensic investigation.



\appendix
\section{Examples of top 10 matches}
In this appendix, additional qualitative results showing examples of top 10 matches returned by the proposed WMM algorithm in EXP3 are presented in Fig. \ref{top10Ranks}.  
\begin{figure*}[p]
\centering
\subfloat[]{\includegraphics[width=6in]{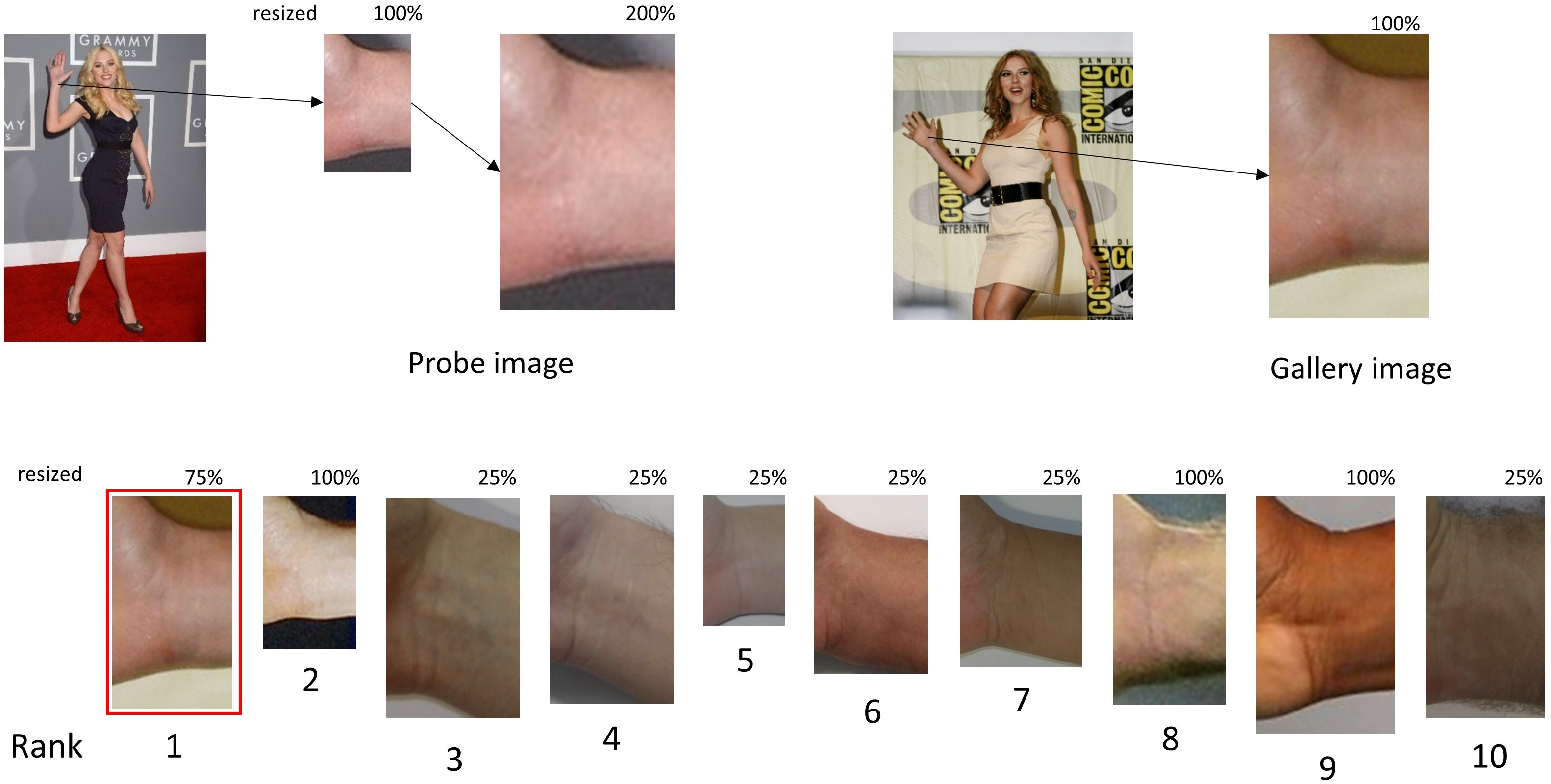}
\label{figTop1}}
\end{figure*}
\begin{figure*}[]
\ContinuedFloat
\subfloat[]{\includegraphics[width=6in]{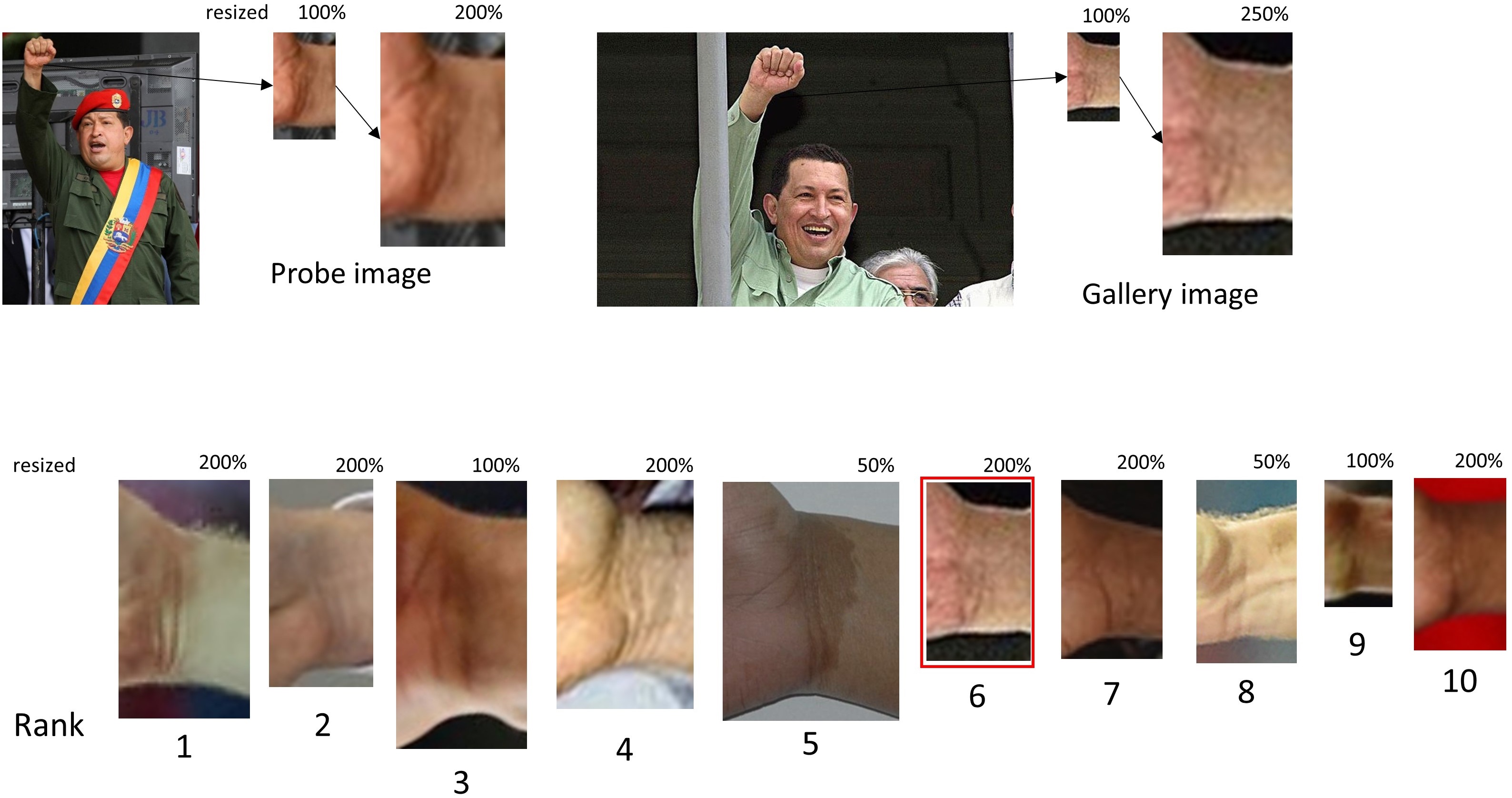}
\label{figTop2}}
\centering
\end{figure*}

\begin{figure*}
\centering
\ContinuedFloat
\subfloat[]{\includegraphics[width=6in]{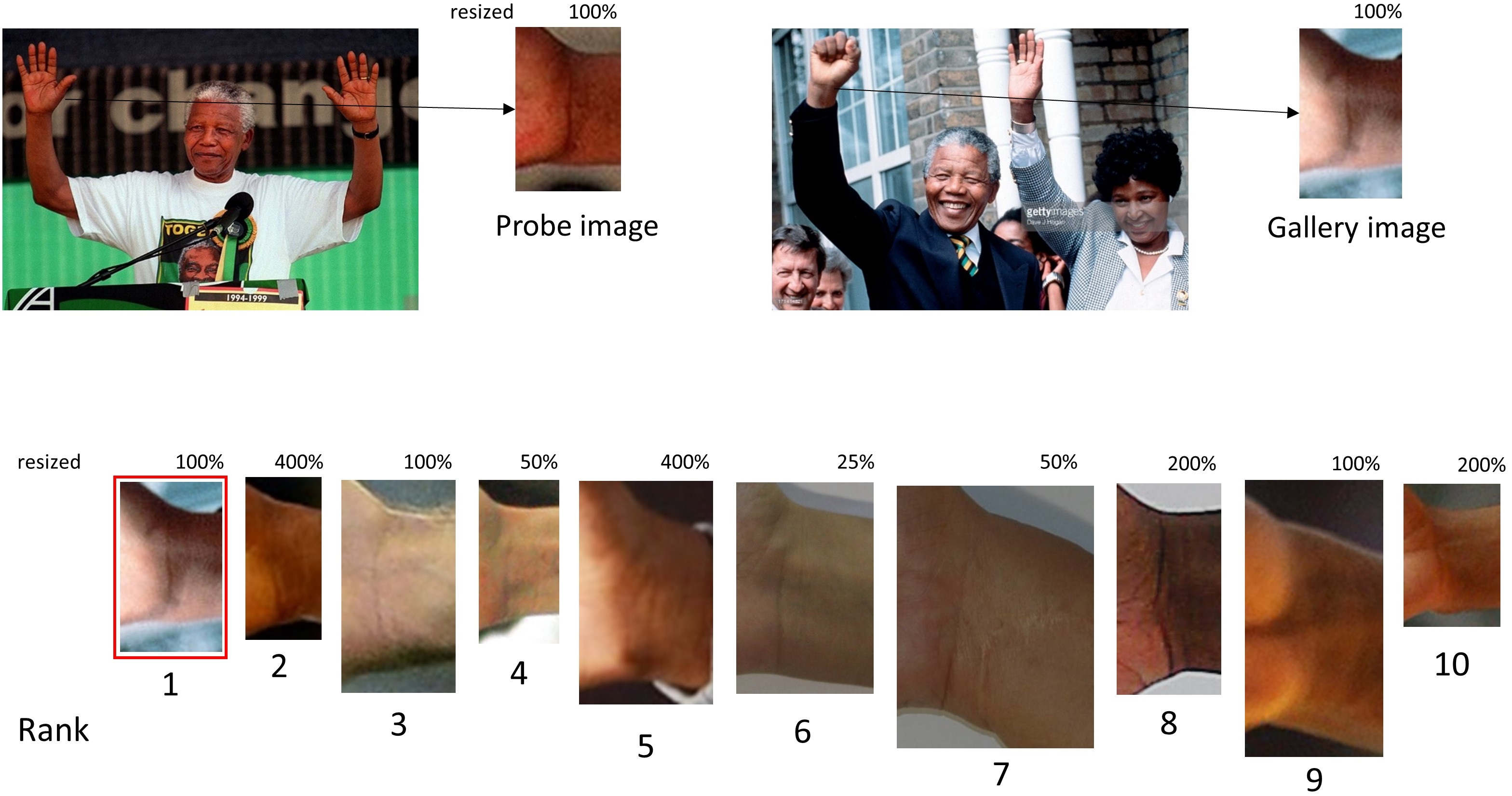}
\label{figTop3}}
\end{figure*}
\begin{figure*}
\ContinuedFloat
\subfloat[]{\includegraphics[width=6in]{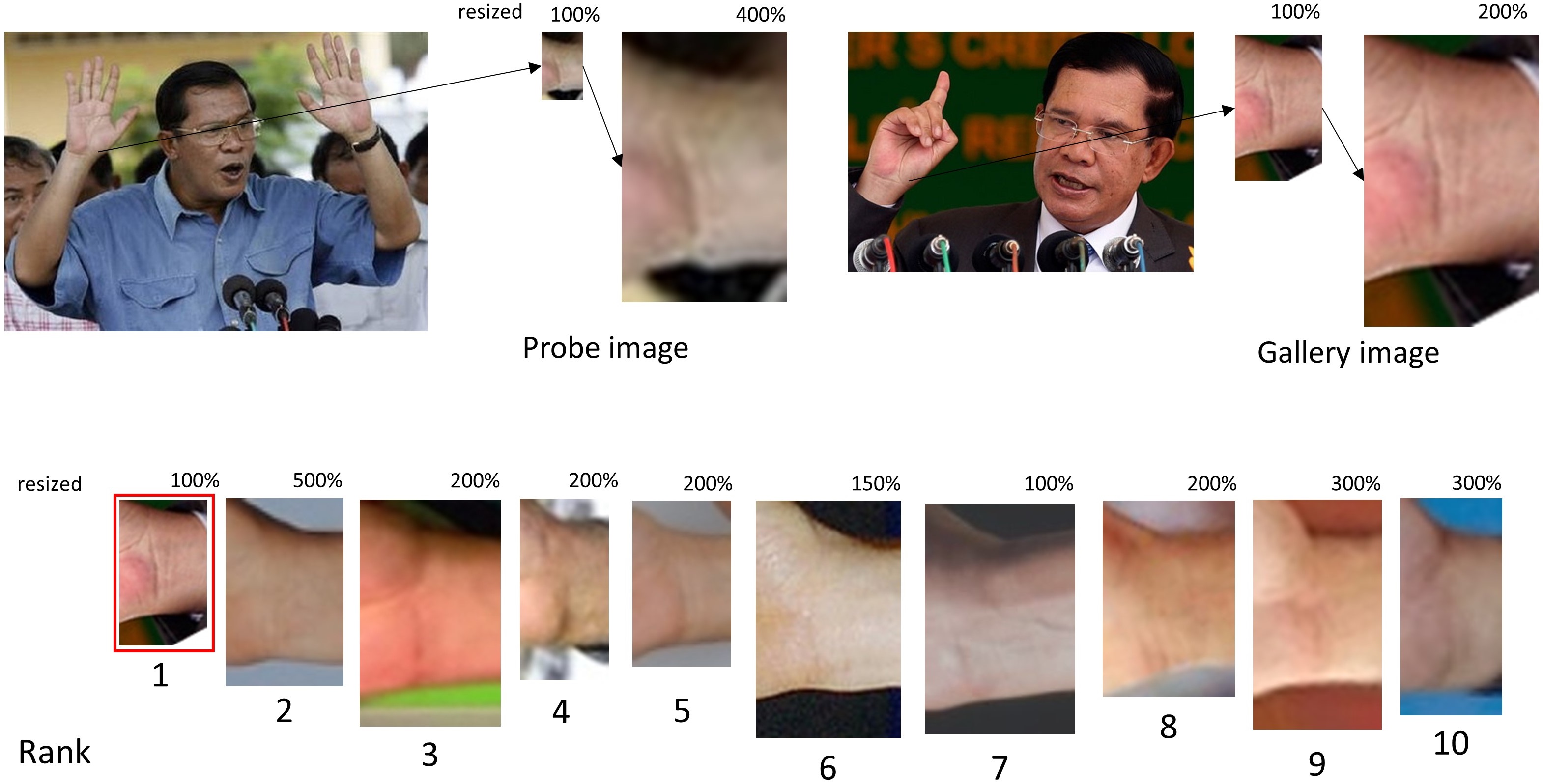}
\label{figTop4}}
\centering
\end{figure*}

\begin{figure*}
\centering
\ContinuedFloat
\subfloat[]{\includegraphics[width=6in]{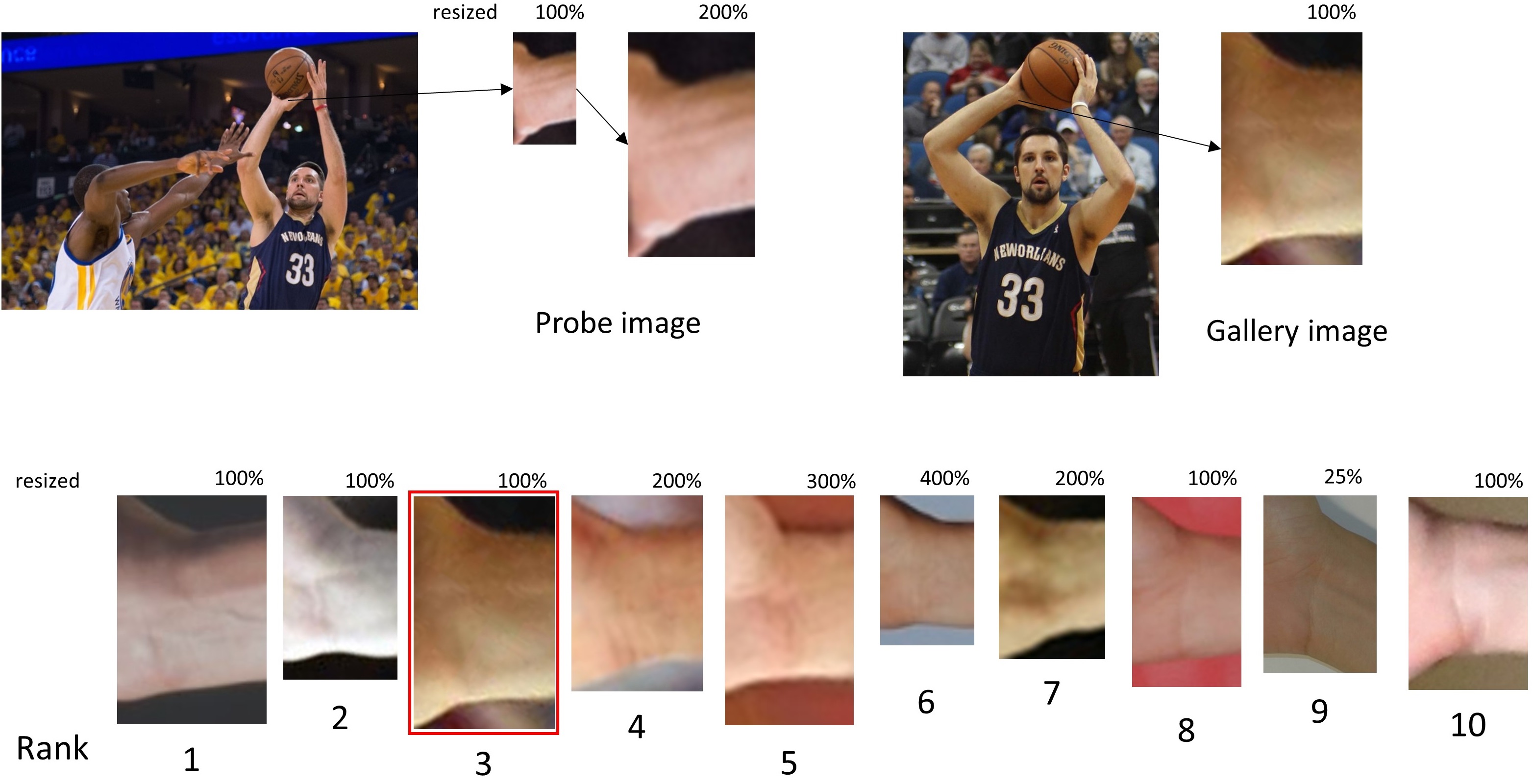}
\label{figTop7}}
\end{figure*}
\begin{figure*}
\ContinuedFloat
\subfloat[]{\includegraphics[width=6in]{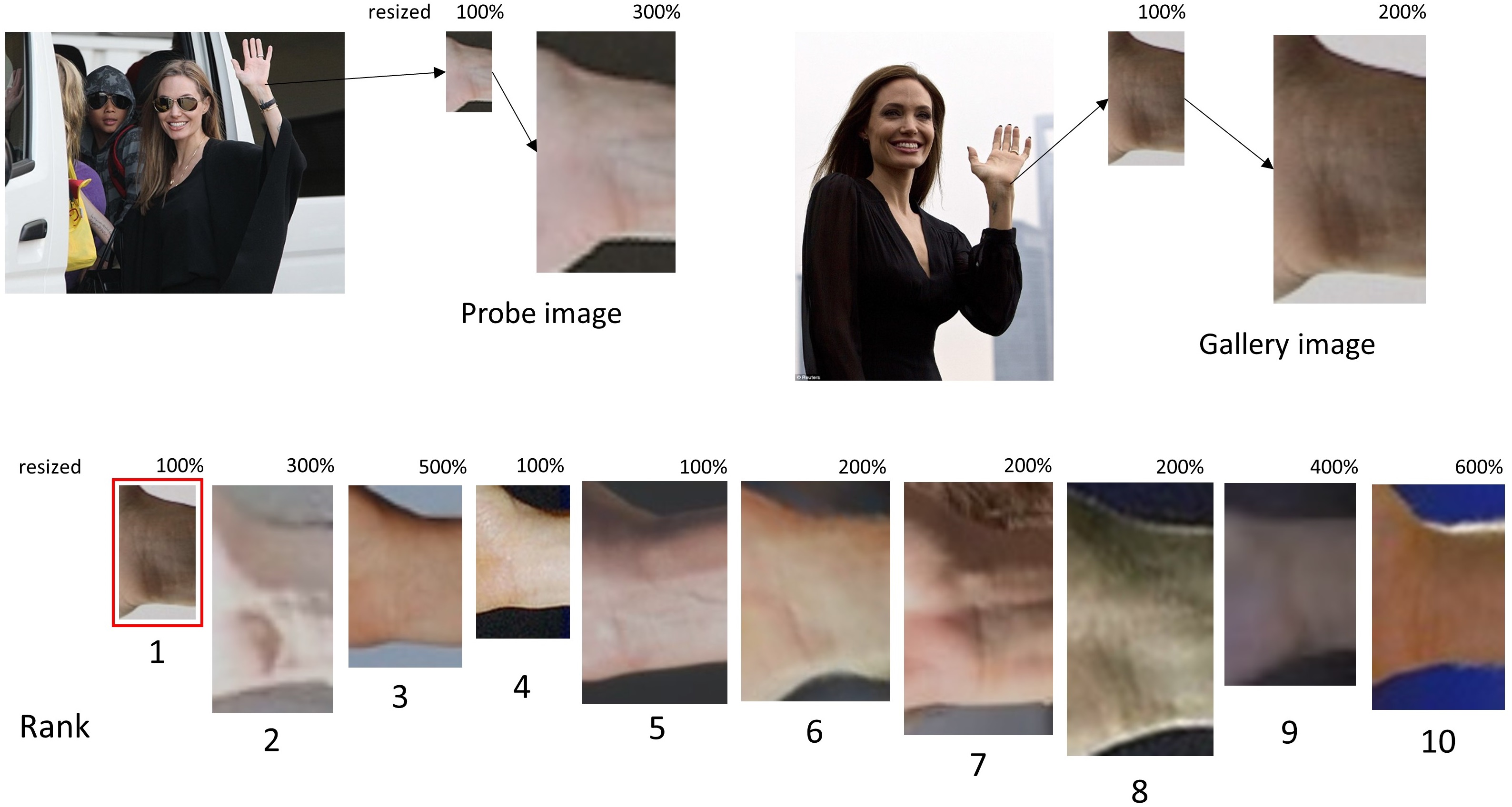}
\label{figTop8}}
\end{figure*}

\begin{figure*}
\centering
\ContinuedFloat
\subfloat[]{\includegraphics[width=6in]{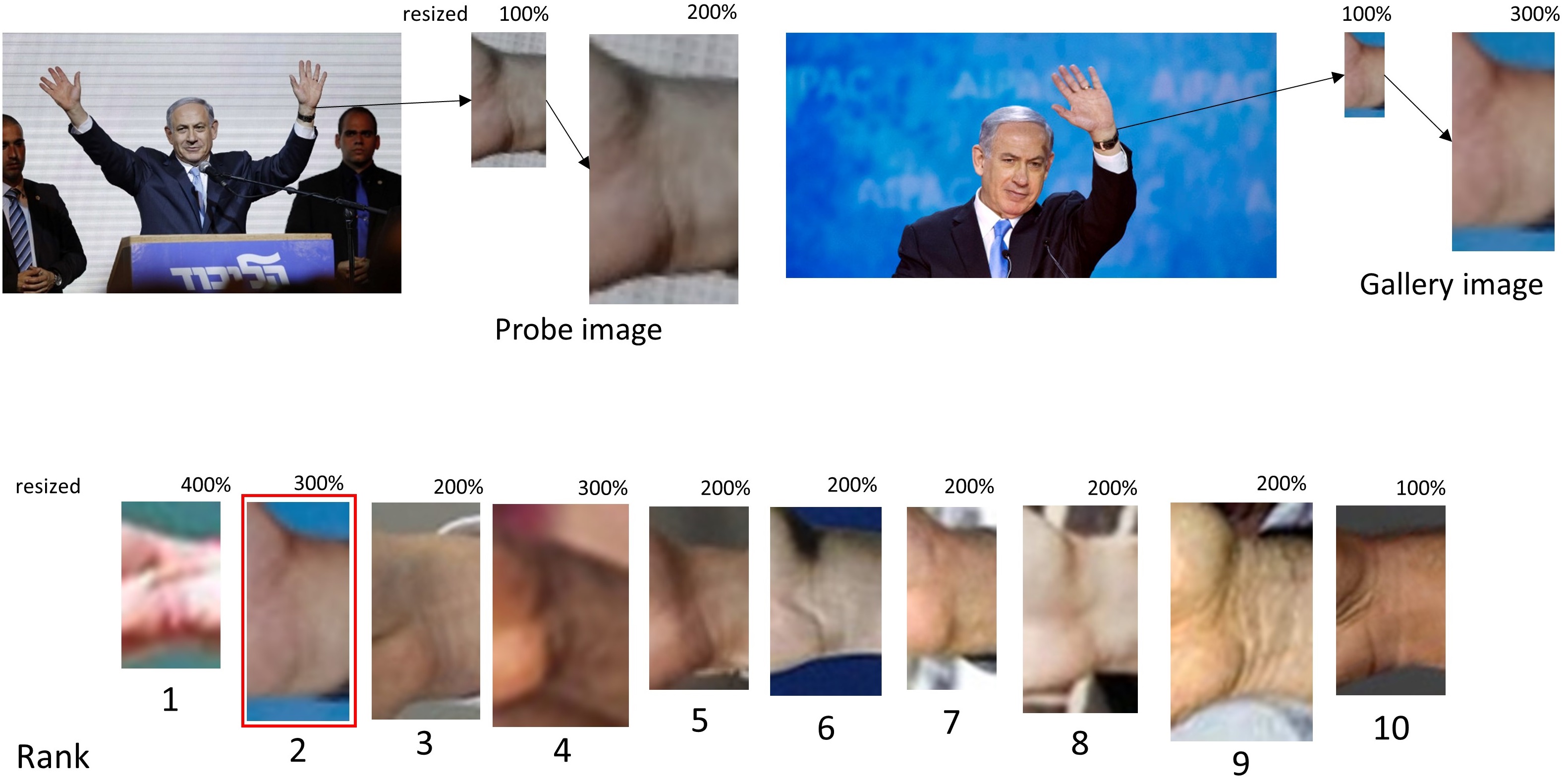}
\label{figTop11}}
\end{figure*}
\begin{figure*}
\setcounter{figure}{15}
\ContinuedFloat
\subfloat[]{\includegraphics[width=6in]{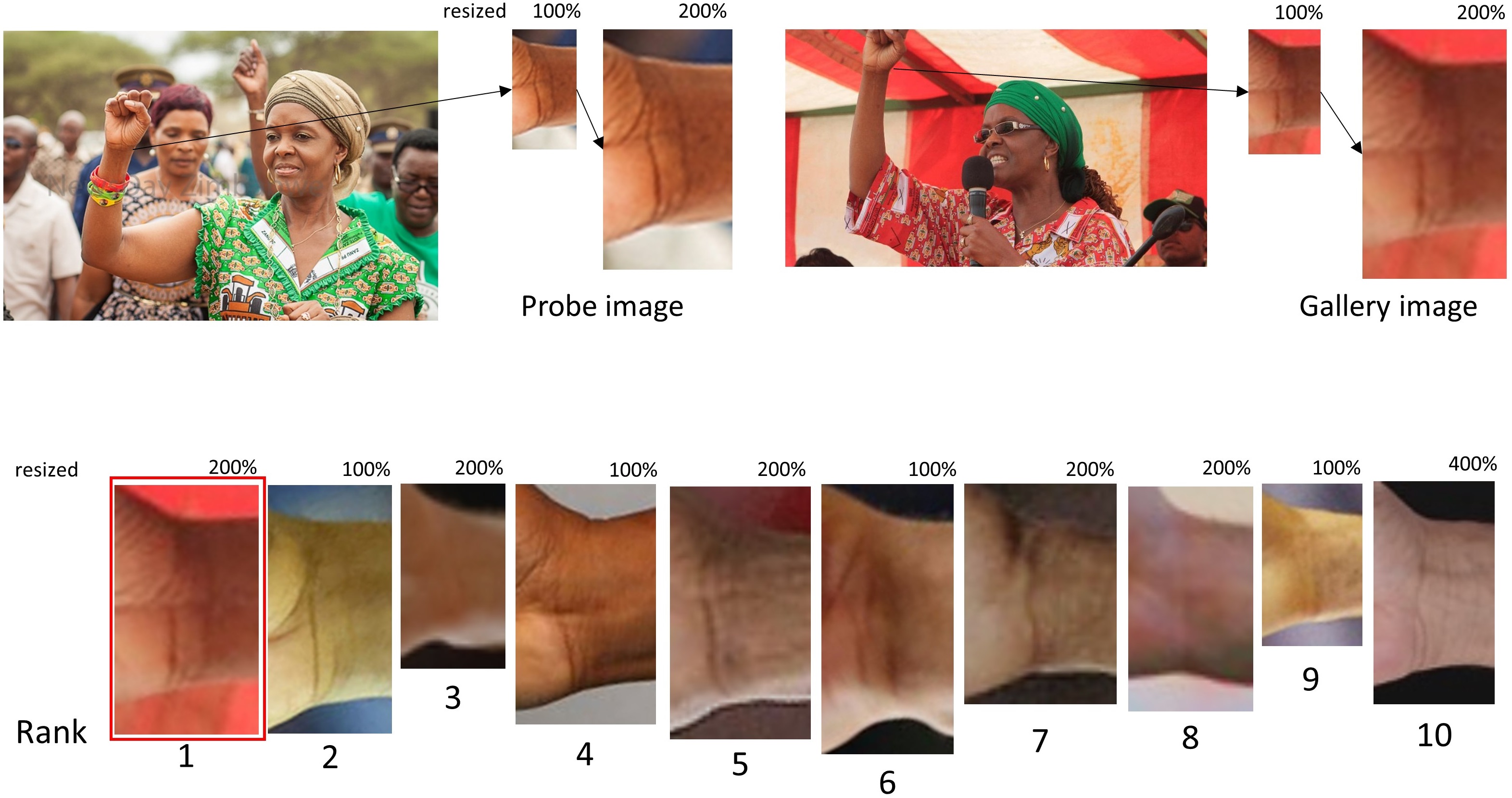}
\label{figTop12}}
\caption{Top 10 matches returned by the WMM algorithm in EXP3.  The genuine matches are highlighted with a red box. Wrist images are resized for better visualization. The scale is indicated above each image.}
\label{top10Ranks}
\end{figure*}


\newpage
\section*{Acknowledgement}
This work is partially supported by the Ministry
of Education, Singapore through Academic Research Fund Tier 2, MOE2016-T2-1-042(S).
\bibliographystyle{elsarticle-num} 

\section*{References}
\bibliography{Wrist}




\end{document}